\documentclass[sigconf]{acmart}
\usepackage{float}
\usepackage{tabularray,varwidth,enumitem}
\usepackage{array}

\usepackage{graphics}
\usepackage{wrapfig}
\usepackage{multirow, makecell}
\usepackage{multicol}
\usepackage{caption}
\usepackage{lipsum} 
\usepackage{subcaption}
\usepackage{mathtools}
\usepackage{bm}
\usepackage{tabularx}
\usepackage{booktabs}
\usepackage[table,xcdraw]{xcolor}
\usepackage{placeins}
\AtBeginDocument{%
  }

\copyrightyear{2025}
\acmYear{2025}
\setcopyright{acmlicensed}\acmConference[MMAsia '25]{ACM Multimedia Asia}{December 9--12, 2025}{Kuala Lumpur, Malaysia} \acmBooktitle{ACM Multimedia Asia (MMAsia '25), December 9--12, 2025, Kuala Lumpur, Malaysia}
\acmDOI{10.1145/3743093.3770950} \acmISBN{979-8-4007-2005-5/2025/12}

\begin{document}

\title{Q-Adapter: Visual Query Adapter\\ for Extracting Textually-related Features in Video Captioning}
\renewcommand{\shorttitle}{Q-Adapter: Visual Adapter for Extracting Textually-related Features in Video Captioning}


\author{Junan Chen}
\orcid{0009-0001-3649-8651}
\affiliation{%
  \institution{Nagoya University}
  \city{Nagoya}
  \state{Aichi}
  \country{Japan}
}
\email{chenj@cs.is.i.nagoya-u.ac.jp}

\author{Trung Thanh Nguyen}
\orcid{0000-0001-8976-2922}
\affiliation{%
  \institution{Nagoya University}
  \city{Nagoya}
  \state{Aichi}
  \country{Japan}
}
\email{nguyent@cs.is.i.nagoya-u.ac.jp}

\author{Takahiro Komamizu}
\orcid{0000-0002-3041-4330}
\affiliation{%
  \institution{Nagoya University}
  \city{Nagoya}
  \state{Aichi}
  \country{Japan}
}
\email{taka-coma@acm.org}

\author{Ichiro Ide}
\orcid{0000-0003-3942-9296}
\affiliation{%
  \institution{Nagoya University}
  \city{Nagoya}
  \state{Aichi}
  \country{Japan}
}
\email{ide@i.nagoya-u.ac.jp}

\renewcommand{\shortauthors}{Junan Chen et al.}

\begin{abstract}
Recent advances in video captioning are driven by large-scale pretrained models, which follow the standard ``pre-training followed by fine-tuning'' paradigm, where the full model is fine-tuned for downstream tasks. Although effective, this approach becomes computationally prohibitive as the model size increases. The Parameter-Efficient Fine-Tuning (PEFT) approach offers a promising alternative, but primarily focuses on the language components of Multimodal Large Language Models (MLLMs). Despite recent progress, PEFT remains underexplored in multimodal tasks and lacks sufficient understanding of visual information during fine-tuning the model. To bridge this gap, we propose \textbf{Q}uery-\textbf{Adapter} (\textbf{Q-Adapter}), a lightweight visual adapter module designed to enhance MLLMs by enabling efficient fine-tuning for the video captioning task. Q-Adapter introduces learnable query tokens and a gating layer into Vision Encoder, enabling effective extraction of sparse, caption-relevant features without relying on external textual supervision. We evaluate Q-Adapter on two well-known video captioning datasets, MSR-VTT and MSVD, where it achieves state-of-the-art performance among the methods that take the PEFT approach across BLEU@4, METEOR, ROUGE-L, and CIDEr metrics. Q-Adapter also achieves competitive performance compared to methods that take the full fine-tuning approach while requiring only 1.4\% of the parameters. We further analyze the impact of key hyperparameters and design choices on fine-tuning effectiveness, providing insights into optimization strategies for adapter-based learning. These results highlight the strong potential of Q-Adapter in balancing caption quality and parameter efficiency, demonstrating its scalability for video--language modeling.
\end{abstract}




\maketitle

\section{Introduction}
\label{sec:introduction}

With the growing demand for advanced video understanding, the video captioning technique is rapidly developing. It aims to generate accurate and contextually relevant textual descriptions of video content. Early studies~\cite{rohrbach2013translating,rohrbach2015dataset} in video captioning primarily employed template-based methods, formulating the task as word prediction at predefined positions within fixed sentence structures. In recent years, the focus has shifted from word-level to sentence-level supervision, with increasing emphasis on end-to-end generation frameworks that align Vision Encoder features with semantic representations~\cite{pan2016jointly}. This alignment transforms video into structured embeddings that Text Decoders can interpret effectively.

To improve cross-modal alignment, several studies~\cite{2020Learning,2020Spatio} have introduced additional supervisory signals to enhance the extraction of semantically relevant visual features. However, the limited size and diversity of publicly available video-caption datasets present significant challenges for effective training. To address this, contrastive learning-based pretraining for image--text alignment~\cite{radford2021learning,li2021align,li2023blip} has been widely adopted, with subsequent task-specific post-training applied to capture temporal dynamics in video~\cite{yu2024eliciting,xu2023mplug,chen2023vast}. While these methods leverage alignment capabilities acquired through contrastive learning, they largely follow the conventional paradigm of ``pre-training followed by fine-tuning'', where full model tuning often leads to challenges such as catastrophic forgetting and suboptimal parameter efficiency~\cite{zhai2023investigating}.


Recently, the development of Multimodal Large Language Models (MLLMs) has introduced new approaches for efficient fine-tuning in vision--language tasks \cite{2024A,huang2024vtimellm,li2024videochatchatcentricvideounderstanding,li2023blip}. 
Methods such as VTimeLLM~\cite{huang2024vtimellm} employ Low-Rank Adaptation (LoRA)~\cite{hu2022lora} to the text encoder with multi-stage training on curated question-answering datasets, showing promising results in dense video captioning.
Other methods incorporate external textual knowledge via Retrieval-Augmented Generation (RAG) to enhance visual understanding~\cite{0Do,2021Open}.
While these methods have demonstrated effectiveness, they often depend heavily on large external text datasets and the reasoning abilities of MLLMs, rather than focusing on improving how the model extracts and utilizes meaningful visual information from videos.
Moreover, they tend to overlook the sparse and uneven distribution of informative content in video, which limits the quality of visual grounding essential for accurate caption generation.

To address these limitations, we introduce an adapter-based fine-tuning method to Vision Encoder of an MLLM by inserting lightweight adapter modules into each layer of the frozen pretrained backbone and updating only these adapter parameters to enhance the visual representation while maintaining the parameter efficiency. In this paper, we propose \textbf{Q}uery-\textbf{Adapter} (\textbf{Q-Adapter}), a novel visual adaptation method for fine-tuning an MLLM in video captioning tasks. 
The proposed method introduces \textit{learnable query tokens} into a lightweight adapter architecture. 
Unlike prior methods that rely heavily on external text sources, Q-Adapter leverages language modeling loss as implicit guidance, allowing the model to dynamically learn which visual features are more relevant for the caption generation. 
By selectively injecting queries into the Vision Encoder, Q-Adapter enables a more effective identification of sparse and semantically informative frames, without requiring explicit annotations or external corpora.

Our key contributions are summarized as follows:
\begin{itemize}
    \item \textbf{Q-Adapter}: We propose a novel query-guided visual adaptation method for efficient fine-tuning of MLLMs in video captioning, introducing learnable query tokens and a gating mechanism into Vision Encoder to selectively extract sparse, caption-relevant features without external textual supervision, enhancing visual grounding under tight parameter constraints.
    \item \textbf{Layer-Wise Adapter Placement}: We systematically evaluate the adapter insertion at progressively deeper layers within Vision Encoder and analyze insertion patterns, providing practical guidelines for PEFT in this context.
    \item \textbf{Effectiveness and Efficiency}: We show that Q-Adapter achieves state-of-the-art performance among methods that take the PEFT approach on the MSR-VTT~\cite{xu2016msr} and MSVD~\cite{chen2011collecting} datasets, using only 1.4\% of the model parameters compared to the full fine-tuning approach, indicating a favorable balance between efficiency and caption quality.
\end{itemize}
The remainder of this paper is structured as follows: Section 2 reviews related work, Section 3 explains the proposed method, Section 4 analyzes the experimental results, and Section 5 concludes the paper and summarizes the main contributions. 

\section{Related Work}
\label{sec:related_work}
Video captioning aims to generate coherent and contextually accurate textual descriptions of visual content by modeling spatio--temporal information in videos. 
Recent advances leverage large-scale pretrained multimodal models such as Contrastive Language-Image Pre-training (CLIP)~\cite{radford2021learning}, shifting the focus toward adapting these models to vision--language tasks. 
Some methods incorporate post-training modules to capture temporal dependencies~\cite{yu2024eliciting, luo2022clip4clip, tang2021clip4caption}, while others fine-tune models directly for caption generation~\cite{xu2023mplug, chen2023vast, yang2023vid2seq}. 
While these methods show strong generalization on benchmarks like MSR-VTT~\cite{xu2016msr} and ActivityNet~\cite{caba2015activitynet}, they often struggle to model object relationships and handle sparse visual cues. 
These limitations hinder their ability to produce detailed and context-specific captions in real-world scenarios. 
To overcome these challenges, we propose a visual adaptation method that enhances pretrained vision--language models by improving the extraction of task-relevant visual features for high-quality video captioning.

Large Language Models (LLMs)~\cite{floridi2020gpt, touvron2023llama} and their multimodal extensions MLLMs~\cite{2024A, huang2024vtimellm, li2024videochatchatcentricvideounderstanding} have shown impressive generative capabilities. 
However, fine-tuning these large models is often computationally expensive, motivating the introduction of the PEFT approach, which aims to reduce trainable parameters in general while maintaining the performance. It falls into two categories: adapter-based methods~\cite{houlsby2019parameter} which insert small learnable modules into Transformer layers, and low-rank methods like LoRA~\cite{hu2022lora} which reparameterize weight matrices to limit parameter updates.
While PEFT has seen success in natural language processing and is increasingly applied to vision tasks~\cite{yangaim, mercea2024time, yin20245}, its application in MLLMs remains limited. 
Existing studies primarily focus on language components~\cite{sung2022vl, zhang2023llama}, leaving the vision encoders largely unmodified. 
Here, we propose a visual adaptation method that extends PEFT to the vision backbone of an MLLM, improving the efficiency of video captioning tasks.

\section{Methodology}
\label{sec:methodology}
\begin{figure*}[t]
    \centering
    \begin{subfigure}[t]{0.58\textwidth}
        \centering
        \includegraphics[width=\textwidth]{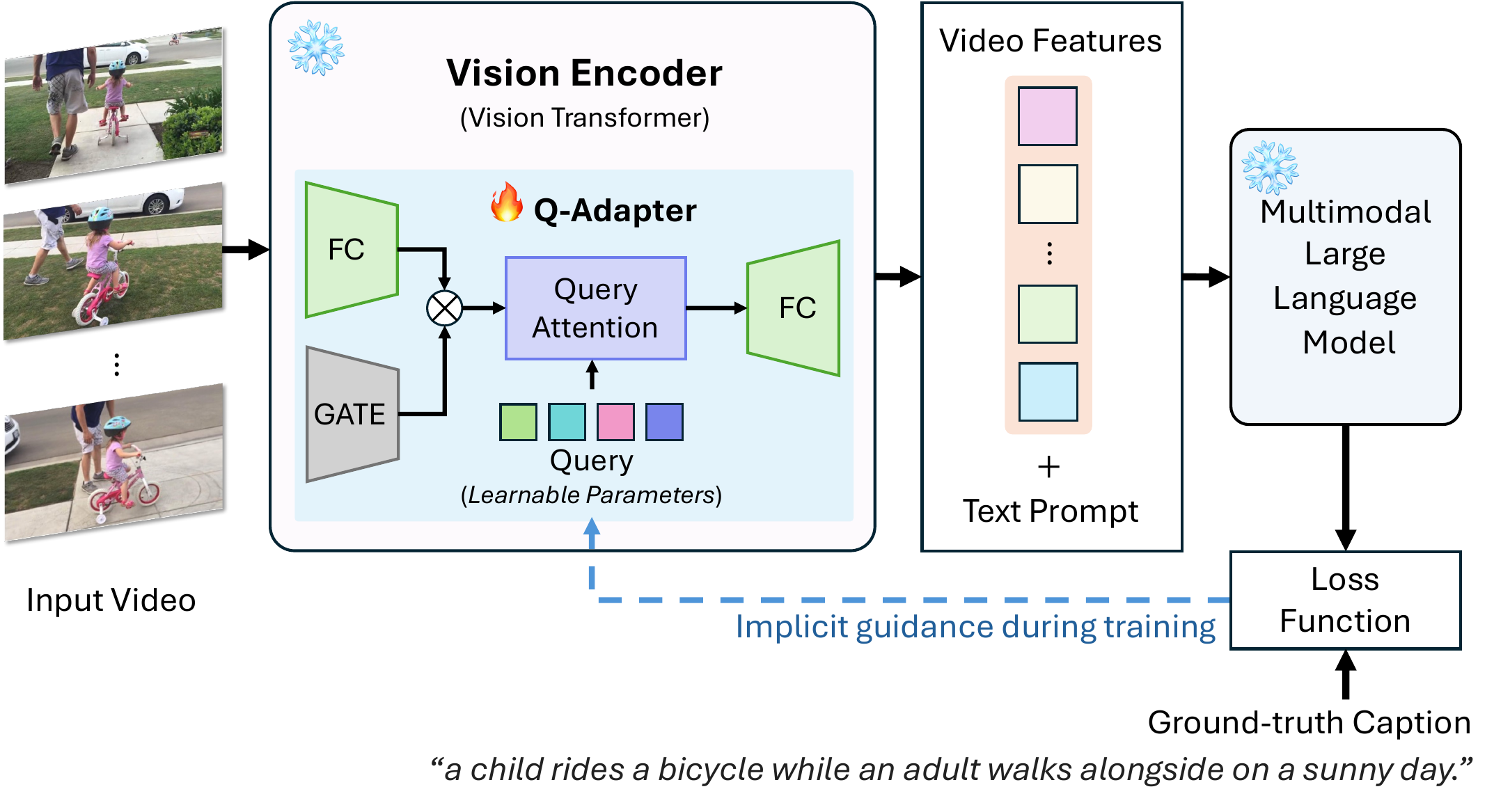}
        \caption{Q-Adapter inserted into Vision Encoder to extract caption-relevant features via query-guided attention and gated layer.}
        \label{fig:overview}
    \end{subfigure} %
    \hfill
    \begin{subfigure}[t]{0.37\textwidth}
        \centering
        \includegraphics[width=\textwidth]{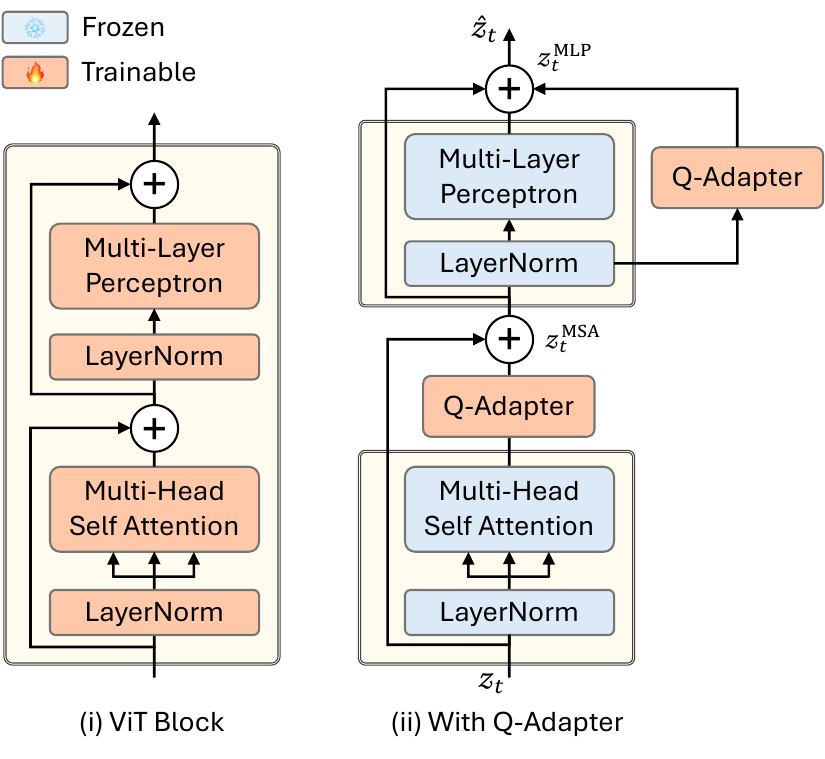}
        \caption{Comparison between (i) standard Vision Transformer (ViT) block and (ii) insertion of Q-Adapter.}
        \label{fig:fig2}
    \end{subfigure}
    \caption{Overview of the video captioning framework and insertion of the proposed Q-Adapter into Vision Encoder.}
    \label{fig:combined}
\end{figure*}

The conventional full fine-tuning approach updates all the parameters of a pretrained model during training. 
In contrast, the adapter fine-tuning~\cite{houlsby2019parameter} approach freezes the pretrained backbone and updates only a small set of newly introduced adapter parameters.
Let $\bm{\phi}_w(\cdot)$ denote a pretrained neural network with parameters $w$, and $\mathcal{D} = \{(\bm{x}_i, \bm{y}_i)\}_{i=1}^N$ the training dataset, the optimization objectives of full and adapter fine-tunings can be formalized as:
\begin{align}
\text{Full fine-tuning:} \quad & \arg\min_{w} \ \mathcal{L}(\bm{\phi}_w(\bm{x}), \bm{y}), \label{eq:full_finetuning} \\
\text{Adapter fine-tuning:} \quad & \arg\min_{\theta} \ \mathcal{L}(\bm{\phi}_{w, \theta}(\bm{x}), \bm{y}), \label{eq:adapter}
\end{align}
where $\bm{x}$ denotes the model input, $\bm{y}$ the corresponding ground-truth label, $\theta$ the adapter parameters, and $\mathcal{L}$ the task-specific loss function. In Eq.~\ref{eq:adapter}, backbone parameters $w$ are frozen while only adapter $\theta$ is updated.

In adapter fine-tuning, the placement of adapter modules is often determined by the characteristics of the downstream task. 
While adapters are typically inserted into intermediate layers of the network, there is no universally optimal configuration. 
Prior studies showed that the effectiveness of adapter placement can vary significantly depending on the task objectives and data modalities~\cite{houlsby2019parameter, ruckle2020adapterdrop, chen2022adaptformer, yangaim}. 
This motivates the need for task-specific design choices when integrating adapters to fine-tune an MLLM.

\subsection{Overview of the Proposed Framework} 
Figure~1(a) shows an overview of the proposed video captioning framework.
Given an input video represented as a sequence of $T$ frames $\mathcal{V} = \{\bm{v}_1, \bm{v}_2, \ldots, \bm{v}_T\}$, the goal of video captioning is to generate a textual description $\bm{\hat{y}} = (\hat{y}_1, \hat{y}_2, \ldots, \hat{y}_L)$, where each $\hat{y}_i$ denotes a token in the output caption of length $L$. 
Each video frame $\bm{v}_t \in \mathbb{R}^{H \times W \times C}$, where $H$, $W$, and $C$ denote the height, width, and number of channels of the input frame, respectively, is processed by Vision Encoder to produce a frame-level representation $\bm{z}_t \in \mathbb{R}^{N \times d}$, where $N$ is the number of visual tokens and $d$ is the embedding dimension. These representations are then temporally aggregated into a unified video-level representation as:
\begin{equation}
\bm{z} = \mathcal{F}(\{\bm{z}_1, \bm{z}_2, \ldots, \bm{z}_T\}),
\end{equation}
where $\mathcal{F}(\cdot)$ denotes a generic temporal fusion function (e.g.,  concatenation, pooling, or self-attention).
The resulting video representation $\bm{z}$ is concatenated with a text prompt $\bm{p}$ and fed into Text Decoder to generate the caption in an autoregressive manner as:
\begin{equation}
\hat{y_i} = \text{Decoder}(\hat{y}_{<i}, [\bm{p}; \bm{z}]), \quad i = \{1, 2, \ldots, L\},
\end{equation}
where $\hat{y}_{<i}$ denotes all previously generated tokens and $[\bm{p}; \bm{z}]$ the concatenation of the text prompt and the fused video features. 
%
The model is trained using Cross Entropy loss, which minimizes the negative log-likelihood of the ground-truth caption~as:
\begin{equation}
\mathcal{L} = - \sum_{i=1}^{L} \log p(\hat{y}_i \mid \hat{y}_{<i}, [\bm{p}; \bm{z}]).
\end{equation}

This formulation highlights that the effectiveness of video captioning largely depends on the quality of the video representation $\bm{z}$, which is obtained from Vision Encoder.
However, standard fine-tuning methods typically focus on Text Decoder and struggle to adapt to the sparse and uneven distribution of informative content in videos. 
To enhance the quality of visual representations and enable efficient adaptation for downstream captioning tasks, we propose Q-Adapter, which is inserted into the Vision Encoder. 
This adapter is designed as a lightweight visual module based on learnable query tokens that selectively extracts caption-relevant features without relying on additional text supervision.

\subsection{Q-Adapter}
Unlike conventional bottleneck adapters that rely solely on projection layers, the proposed Q-Adapter incorporates a learnable query-attention mechanism and a gating network to enhance its ability to extract sparse and semantically rich visual features.

\subsubsection{Query-Attention Mechanism}
For each input video frame $t \in \{1, \ldots ,T\}$, we denote the encoded feature map as \( \bm{z}_t \in \mathbb{R}^{N \times d} \), where \( N \) is the number of visual tokens and \( d \) is the embedding dimension. To enhance the discriminative capacity of these features, Q-Adapter introduces a set of learnable query tokens \( \bm{q} \in \mathbb{R}^{M \times d} \), where \( M \) is the number of queries. These tokens are designed to extract task-relevant information from \( \bm{z}_t \) via cross-attention.
We follow the standard scaled dot-product attention mechanism, where queries \( Q \), keys \( K \), and values \( V \) are defined~as:
\begin{equation}
Q = \bm{q} \text{W}^Q, \quad K = \bm{z}_t \text{W}^K~\text{and} \quad V = \bm{z}_t \text{W}^V,
\end{equation}
where \( \text{W}^Q, \text{W}^K,~\text{and}~\text{W}^V \in \mathbb{R}^{d \times d} \) are learned projection matrices. The query-guided attention output is then computed~as:
\begin{equation}
\bm{o}_t = \text{LayerNorm} \left( \text{softmax} \left( \frac{Q K^\top}{\sqrt{d}} \right) V \right),
\end{equation}
where \( \bm{o}_t \in \mathbb{R}^{M \times d} \) is the attended representation that captures the most relevant information from \( \bm{z}_t \) with respect to each query token.
To preserve the spatial structure required for subsequent temporal fusion, the output is projected back to the original token length via a learned transformation \( \Psi: \mathbb{R}^{M \times d} \rightarrow \mathbb{R}^{N \times d} \), followed by residual addition as:
\begin{equation}
\bm{z}_t^{\text{q}} = \Psi(\bm{o}_t) + \bm{z}_t.
\end{equation}

This formulation allows Q-Adapter to act as a content-aware refinement layer, selectively attending to semantically important regions while maintaining the computational efficiency and the structural alignment with the original token space.

\subsubsection{Gated Feature Layer}
To account for the potential distribution mismatch caused by freezing the pretrained encoder weights, we introduce a gating mechanism that adaptively modulates the input of the adapter. 
Specifically, we apply a Fully Connected (FC) layer to project the input feature \( \bm{z}_t \in \mathbb{R}^{N \times d} \), followed by element-wise multiplication with a learnable gating layer. 
The gated output is computed as:
\begin{equation}
\bm{z}_t^{\text{g}} = \text{Gate}(\bm{z}_t) \circ \text{FC}(\bm{z}_t),
\end{equation}
where \( \text{FC}(\cdot): \mathbb{R}^{N \times d} \rightarrow \mathbb{R}^{N \times d'} \) is a dimensionality-reducing linear transformation, \( \text{Gate}(\cdot) \) is a learnable gating layer implemented as a multi-layer perceptron, and \( \circ \) denotes element-wise multiplication. 
The gated representation $\bm{z}_t^{\text{g}}$ is used as the Key (K) and Value (V) inputs in the query-attention mechanism.

\subsubsection{Insertion into Vision Encoder}
As illustrated in Fig.~1(b), we insert Q-Adapter into each block of Vision Encoder to enable PEFT. Here, spatial and temporal features are fused before being processed by Vision Transformer (ViT)~\cite{vit2020} for improved spatio-temporal modeling. Specifically, we apply Q-Adapter to each ViT block: one for the Multi-head Self-Attention (MSA) layer and another for the Multi-Layer Perceptron (MLP) layer.
This design preserves the original architecture while introducing a lightweight adaptation path that refines visual features without modifying pretrained backbone parameters.
Let \( \bm{z}_t \in \mathbb{R}^{N \times d} \) be the input token sequence to the block, the processing steps are defined~as: 
{
\begin{align}
\bm{z}_t^{\text{MSA}} &= \bm{z}_t + \text{Q-Adapter}(\text{MSA}(\text{LayerNorm}(\bm{z}_t))), \\
\bm{z}_t^{\text{MLP}} &= \bm{z}_t^{\text{MSA}} + \text{MLP}(\text{LayerNorm}(\bm{z}_t^{\text{MSA}})) \notag \\
                      &\quad + \text{Q-Adapter}(\text{LayerNorm}(\bm{z}_t^{\text{MSA}})),  \\
\mathllap{\hat{\bm{z}}_t} &= \bm{z}_t^{\text{MLP}}, 
\end{align}
}

\noindent where \(\bm{z}^{\text{MSA}}\) is the output sequence of the MSA layer, 
and \(\bm{z}^{\text{MLP}}\) the output sequence of the MLP layer. In this structure, only the parameters of Q-Adapter are updated during training, while the original components of Vision Encoder remain frozen. 
This selective tuning allows efficient adaptation to video captioning tasks with minimal additional parameters.

\section{Experiments}
\label{sec:performance_evaluation}
\begin{table*}[t]
\centering
\caption{Comparison of the proposed Q-Adapter with other methods on the MSR-VTT~\cite{xu2016msr} and MSVD~\cite{chen2011collecting} datasets. 
Best and second-best scores for each Fine-Tuning (FT) Approach are highlighted with \textbf{bold} and \underline{underlined} texts, respectively.
}
{
\begin{tabular}{l|c|c|*{4}c|*{4}c}
\toprule
\multirowcell{2}[-2pt][l]{Method} & \multirowcell{2}[-2pt][c]{FT Approach} & \multirowcell{2}[-2pt][c]{FT Ratio~$\downarrow$} & \multicolumn{4}{c|}{MSR-VTT Dataset~\cite{xu2016msr}} & \multicolumn{4}{c}{MSVD Dataset~\cite{chen2011collecting}} \\ \cmidrule(lr){4-7} \cmidrule(lr){8-11}
 &  &  & B@4~$\uparrow$ & M~$\uparrow$ & R-L~$\uparrow$ & C~$\uparrow$ & B@4~$\uparrow$ & M~$\uparrow$ & R-L~$\uparrow$ & C~$\uparrow$  \\ 
\midrule
SwinBERT~\cite{lin2022swinbert} & \multirowcell{5}[0pt][c]{Full} & \multirowcell{5}[0pt][c]{100\%} & 41.90 & 29.90 & 62.10 & 53.80 & 58.20 & 41.30 & 77.50 & 120.60 \\
MV-GPT~\cite{seo2022end} & & & 48.92 & \textbf{38.66} & 64.00 & 60.00 & --- & --- & --- & --- \\
HiTeA~\cite{ye2023hitea} & & & 49.20 & 30.70 & 65.00 & 65.10 & \underline{71.00} & \underline{45.30} & 81.40 & 146.90 \\
mPLUG-2\textsubscript{Base}~\cite{xu2023mplug} &  &  & \underline{52.20} & 32.10 & \underline{66.90} & \underline{72.40} & 69.30 & 45.10 & \underline{81.90} & \underline{148.20} \\
mPLUG-2~\cite{xu2023mplug} &  &  & \textbf{57.80} & \underline{34.90} & \textbf{{70.10}} & \textbf{{80.30}} & \textbf{75.00} & \textbf{48.40} & \textbf{85.30} & \textbf{165.80} \\
\midrule
Qwen2.5-VL~\cite{qwen2.5-VL} & Zero-shot & 0.0\% &  \phantom{0}5.21 & 18.43 & 22.36 & \phantom{0}0.26 & \phantom{0}5.21 & 18.86 & 23.36 & \phantom{00}0.90\\ \cmidrule(lr){1-11}
~\cite{qwen2.5-VL} + LoRA (Attn)~\cite{hu2022lora} & \multirowcell{5}[0pt][c]{PEFT} & \textbf{0.8\%} & 44.22  & 31.58 & 60.33 & 62.55 & \underline{64.69} & \textbf{44.87} & \textbf{80.99} & \underline{139.17}\\
~\cite{qwen2.5-VL} + LoRA (MLP)~\cite{hu2022lora} &  & 1.5\% & 45.14 & 32.03 & 60.88 & 63.10 & 61.58 & 44.05 & 80.45 & 135.67 \\
~\cite{qwen2.5-VL} + LoRA (Attn\&MLP)~\cite{hu2022lora} &  & 2.3\% & 44.78 & 31.71 & 60.48 & 62.36 & 61.71 & 43.28 & 79.45 & 135.72\\
~\cite{qwen2.5-VL} + Adapter~\cite{houlsby2019parameter} & & 2.1\% & \underline{49.52} & \underline{32.21} & \textbf{63.73} & \underline{66.45} & 61.84 & 44.51 & 75.60 & 138.71\\
\rowcolor[HTML]{e5e5e5} ~\cite{qwen2.5-VL} + Q-Adapter (Proposed) &  & \underline{1.4\%} & \textbf{49.84} & \textbf{32.32} & \underline{62.69} & \textbf{67.38} & \textbf{64.84} & \underline{44.55} & \underline{80.78} & \textbf{139.70}\\ 
\bottomrule
\end{tabular}
}
\label{tab:msrvtt_comparison}
\end{table*}

In this section, we evaluate the proposed Q-Adapter on two widely used video captioning datasets and present detailed analysis and discussion.

\subsection{Experimental Settings}
\subsubsection{Datasets}
We conduct experiments on MSR-VTT~\cite{xu2016msr} and MSVD~\cite{chen2011collecting} datasets. 
MSR-VTT consists of 10k YouTube videos (about 10 seconds each) annotated with 200k captions, while MSVD consists of 1,970 YouTube videos (about 10 seconds each) with 120k captions. 
Following prior work~\cite{luo2022clip4clip,xu2023mplug}, we adopt the standard splits: 9k videos for training and 1k for testing on MSR-VTT, and 1k videos for training and 670 for testing on MSVD.

\subsubsection{Implementation Details}
We adopt Qwen2.5-VL-3B\footnote{\url{https://huggingface.co/Qwen/Qwen2.5-VL-3B-Instruct/} (Accessed: 2025-10-10)} as the pretrained backbone for Vision Encoder and Text Decoder. The input resolution of each video frame is standardized to $224 \times 224$ pixels. Following Xu et al.~\cite{xu2023mplug}, we randomly sample 16 frames per video during training and apply uniform sampling during inference to ensure stability and reproducibility. 
All experiments are trained for 10 epochs with a batch size of 8 using AdamW~\cite{loshchilov2017decoupled}, a learning rate of $2 \times 10^{-4}$, and a linear scheduler with a warmup ratio of 0.1. 
    For the proposed method, we explore different numbers of query tokens (\#Queries), ranging from 16 to 224.  For the quantitative results in Section~4.2, we set \#Queries to 16, guided by our ablation results. Q-Adapter is inserted at various layers of Vision Encoder, from the 1st to the 32nd layer, where the 32nd corresponds to the final layer of the 32-layer ViT. 
The text prompt is set as \textit{``What is shown in this video?''}, along with the default system prompt. 
All experiments are conducted on a PC with a single NVIDIA A6000 GPU.

\subsubsection{Evaluation Metrics}
All methods are evaluated on the video captioning task using four standard metrics: BLEU@4 (B@4)~\cite{papineni2002bleu}, METEOR (M)~\cite{banerjee2005meteor}, ROUGE-L (R-L)~\cite{lin2004rouge}, and CIDEr (C)~\cite{vedantam2015cider}. 
These metrics assess the quality of generated captions in term of their similarity to ground-truth references.
To evaluate the training efficiency of methods that take the PEFT approach, we define the Fine-Tuned Ratio (FT Ratio), which highlights the distinction between the PEFT and the full fine-tuning approach as:
\begin{equation} 
 \text{FT Ratio} = \frac{\text{Number of trainable parameters}}{\text{Total number of model parameters}}.
\end{equation}


\subsubsection{Comparison Methods}
We compare the proposed method with state-of-the-art methods that take the full fine-tuning and PEFT approaches in a unified evaluation setting.

\noindent\textbf{Full Fine-Tuning Methods.} 
{SwinBERT}~\cite{lin2022swinbert} adopts an end-to-end sparse attention model with a Swin Transformer~\cite{liu2021swinT} backbone.
{MV-GPT}~\cite{seo2022end} adopts a joint video--text pretraining framework with utterance prediction.
{HiTeA}~\cite{ye2023hitea} adopts a hierarchical temporal-aware model.
{mPLUG-2}~\cite{xu2023mplug} adopts a unified framework with dual Vision Encoders and shared multimodal modules.

\noindent\textbf{PEFT Methods.} 
{Adapter}~\cite{houlsby2019parameter} inserts lightweight adaptive layers after the MSA and MLP modules in each ViT block, which employs a gating mechanism to improve training stability. 
{LoRA}~\cite{hu2022lora} introduces trainable low-rank matrices in parallel with the original pre-trained weights. We explore three integration strategies for LoRA: (1) {LoRA (Attn)} which applies LoRA to the attention projections in the MSA layers, (2) {LoRA (MLP)} which applies LoRA to the feedforward MLP layers, and (3) {LoRA (Attn\&MLP)} which applies LoRA to both the MSA and MLP layers.

\begin{figure*}[t]
    \centering
    \begin{subfigure}{0.245\textwidth}
        \centering
        \includegraphics[width=\linewidth]{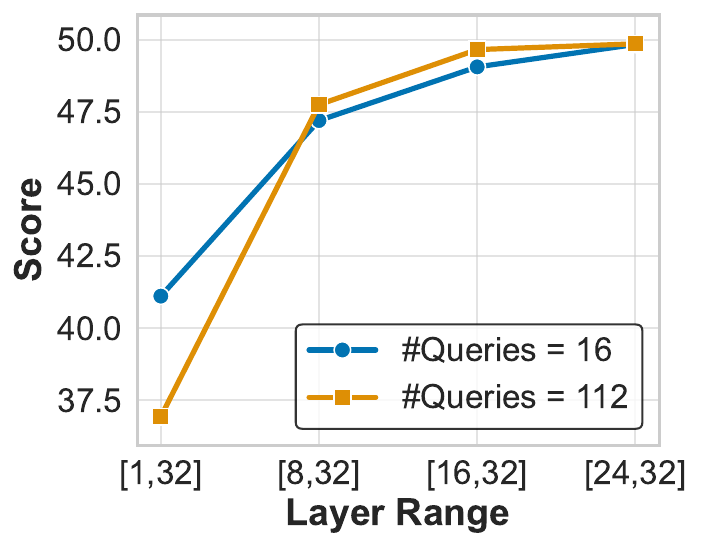}
        \caption{BLEU@4}
    \end{subfigure}
    \begin{subfigure}{0.245\textwidth}
        \centering
        \includegraphics[width=\linewidth]{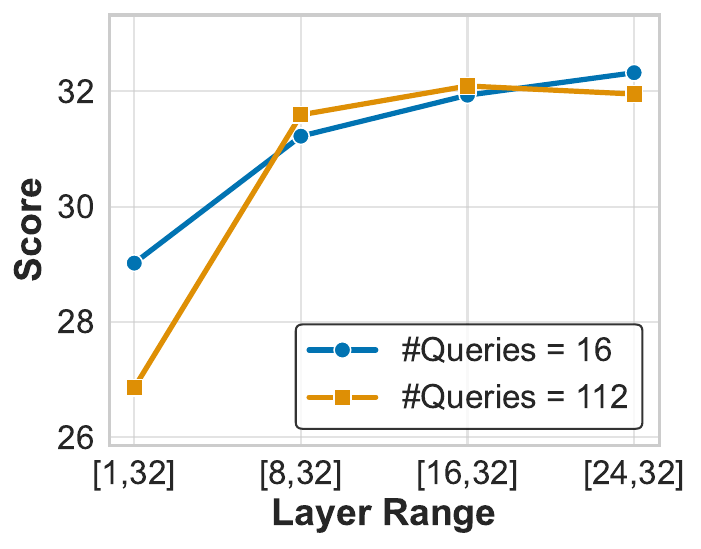}
        \caption{METEOR}
    \end{subfigure}
    \begin{subfigure}{0.245\textwidth}
        \centering
        \includegraphics[width=\linewidth]{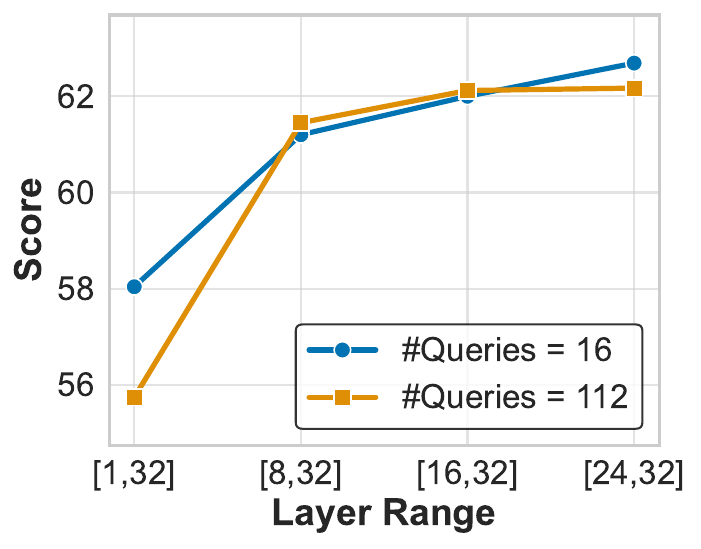}
        \caption{ROUGE-L}
    \end{subfigure}
    \begin{subfigure}{0.245\textwidth}
        \centering
        \includegraphics[width=\linewidth]{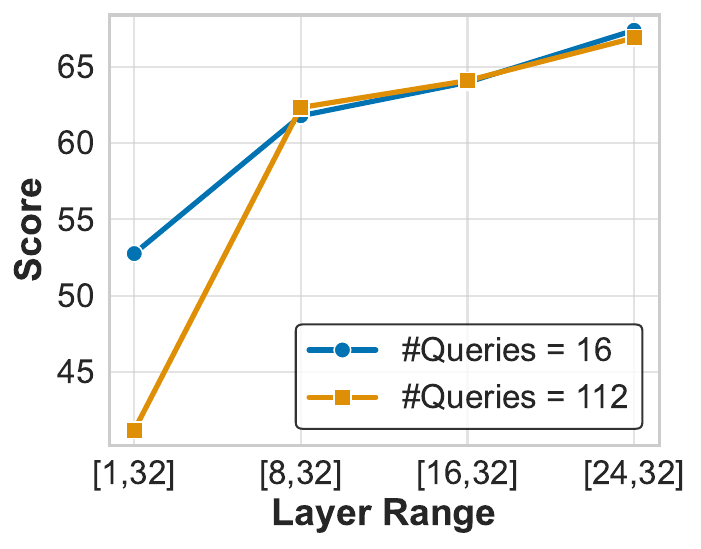}
        \caption{CIDEr}
    \end{subfigure}
    \caption{Comparison across different layer insertion ranges on the MSR-VTT~\cite{xu2016msr} dataset ({\#Queries} = 16 and 112).}
    \label{fig:layer_comparison}
\end{figure*}

\subsection{Quantitative Results}
As shown in Table~\ref{tab:msrvtt_comparison}, the proposed Q-Adapter method achieved strong performance across both the MSR-VTT~\cite{xu2016msr} and MSVD~\cite{chen2011collecting} datasets. 
Compared to the full fine-tuning approach, Q-Adapter outperformed SwinBERT~\cite{lin2022swinbert}, HiTeA~\cite{ye2023hitea}, and MV-GPT~\cite{seo2022end} across most metrics for MSR-VTT, and surpassed SwinBERT~\cite{lin2022swinbert} on all metrics for MSVD, showcasing its competitive effectiveness with only a small fraction of the parameters tuned.
Although mPLUG-2~\cite{xu2023mplug} achieved the highest scores overall due to its larger capacity and full fine-tuning approach, the PEFT approach offered a more efficient alternative, reaching comparable performance while tuning less than 2.3\% of the parameters. 

Among the PEFT methods, Q-Adapter showed a strong and consistent performance with a low FT Ratio (1.4\%).
For MSR-VTT, it achieved the best scores on BLEU@4~\cite{papineni2002bleu}, METEOR~\cite{banerjee2005meteor} and CIDEr~\cite{vedantam2015cider}, and ranked second on ROUGE-L~\cite{lin2004rouge}, slightly behind the standard Adapter~\cite{houlsby2019parameter}, which required a higher FT Ratio (2.1\%). In contrast, LoRA-based methods, while efficient in terms of the number of parameters, demonstrated suboptimal performance and failed to match Q-Adapter in both accuracy and consistency. 
For MSVD, Q-Adapter outperformed the other PEFT methods in BLEU@4 and CIDEr. While LoRA (Attn) slightly exceeded Q-Adapter in METEOR and ROUGE-L with a marginally lower FT Ratio (0.8\%), Q-Adapter delivered a more balanced performance. 
These results validated the scalability of Q-Adapter and its ability to adapt well across datasets of varying sizes without sacrificing efficiency.

To further analyze the behavior of Q-Adapter, we evaluate its performance under different Vision Encoder layer insertion ranges and query token settings, as illustrated in Fig.~\ref{fig:layer_comparison}. Across all metrics, inserting Q-Adapter into only deeper layers (e.g., [24,\,32]) consistently yielded better results than inserting it from early layers (e.g., [1,\,32] or [8,\,32]). This trend held for both small and large numbers of query tokens. These findings suggest that deeper layers, which encode higher-level semantic features, are more critical for vision--language alignment tasks. 
By selectively fine-tuning these layers, Q-Adapter achieved a better trade-off between performance and efficiency.

\subsection{Ablation Studies}

\paragraph{Impact of number of query tokens in Q-Adapter}
We investigate the effect of the number of query tokens (\#Queries) in the proposed Q-Adapter method by evaluating four configurations: 16, 64, 112, and 224. 
As shown in Table~\ref{tab:ablation_query_tokens}, using 16 query tokens yielded the best overall performance across most metrics while maintaining the lowest FT Ratio (1.4\%). 
While increasing the number of queries slightly improved CIDEr~\cite{vedantam2015cider} at 224 queries, the overall gains were marginal and came at the cost of higher parameter usage. 
We attribute strong performance at lower query counts to more focused and efficient attention, enabling each Q-Adapter to extract distinct and task-relevant features without redundancy. 
\paragraph{Effect of Q-Adapter placement in the Vision Encoder}
We conduct experiments to assess the effect of different Q-Adapter placements within Vision Encoder. 
Specifically, we evaluate three placements: (1) {Sequential} where Q-Adapter is inserted after the MSA and MLP layers in each ViT block, (2) {Parallel-MLP} where Q-Adapter is inserted in parallel only within the MLP layer, and (3) {Proposed} where Q-Adapter is placed sequentially after the MSA layer and in parallel within the MLP layer.
As shown in Table~\ref{tab:ablation_placement}, Proposed achieved the highest performance while maintaining a moderate FT Ratio (1.4\%). 
Although Parallel-MLP slightly outperformed Proposed in BLEU@4 and had the lowest FT Ratio (0.7\%), it fell behind on other key metrics. 
Sequential, despite having the same FT Ratio as Proposed, yielded the lowest scores across all metrics.
By targeting different parts of the ViT block, Q-Adapter was better positioned to capture fine-grained contextual dependencies and higher-level semantic information.
\subsection{Qualitative Results}
\begin{figure}[t]
  \centering
  \begin{subfigure}{0.157\textwidth}\hspace{0.01\textwidth}
    \includegraphics[width=\linewidth]{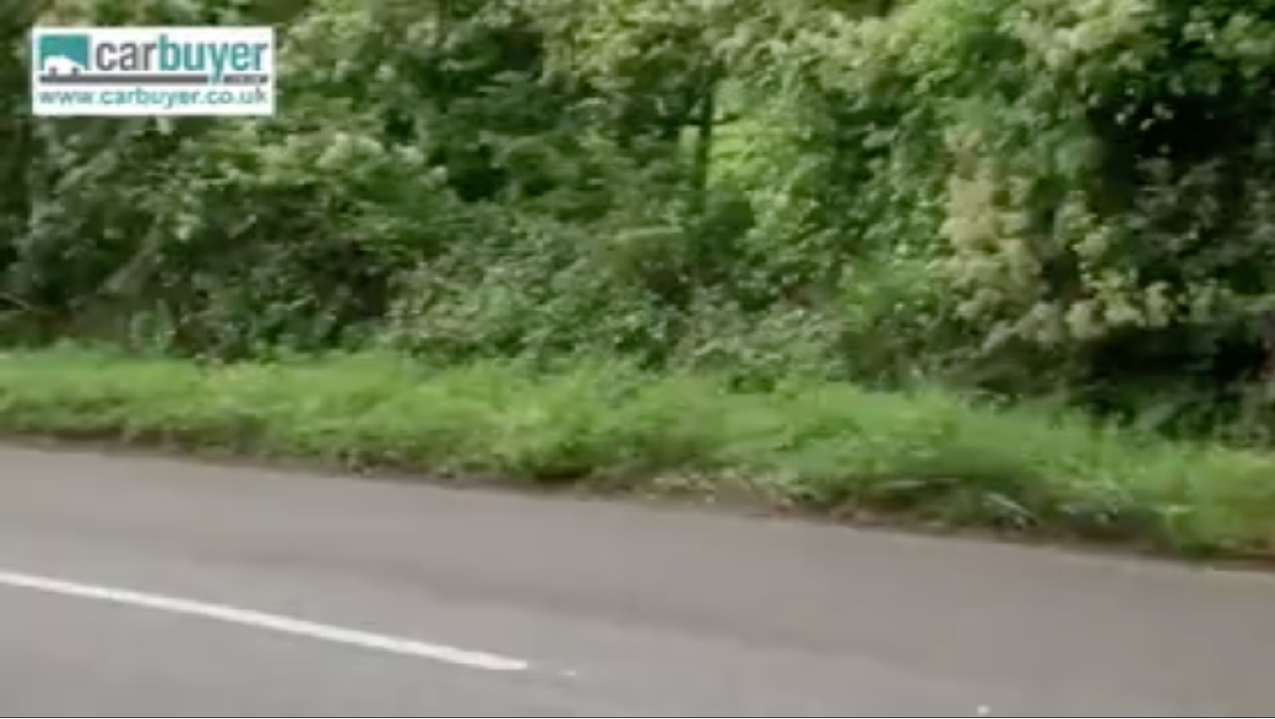}
  \end{subfigure}\hfill
  \begin{subfigure}{0.157\textwidth}\hspace{0.01\textwidth}
    \includegraphics[width=\linewidth]{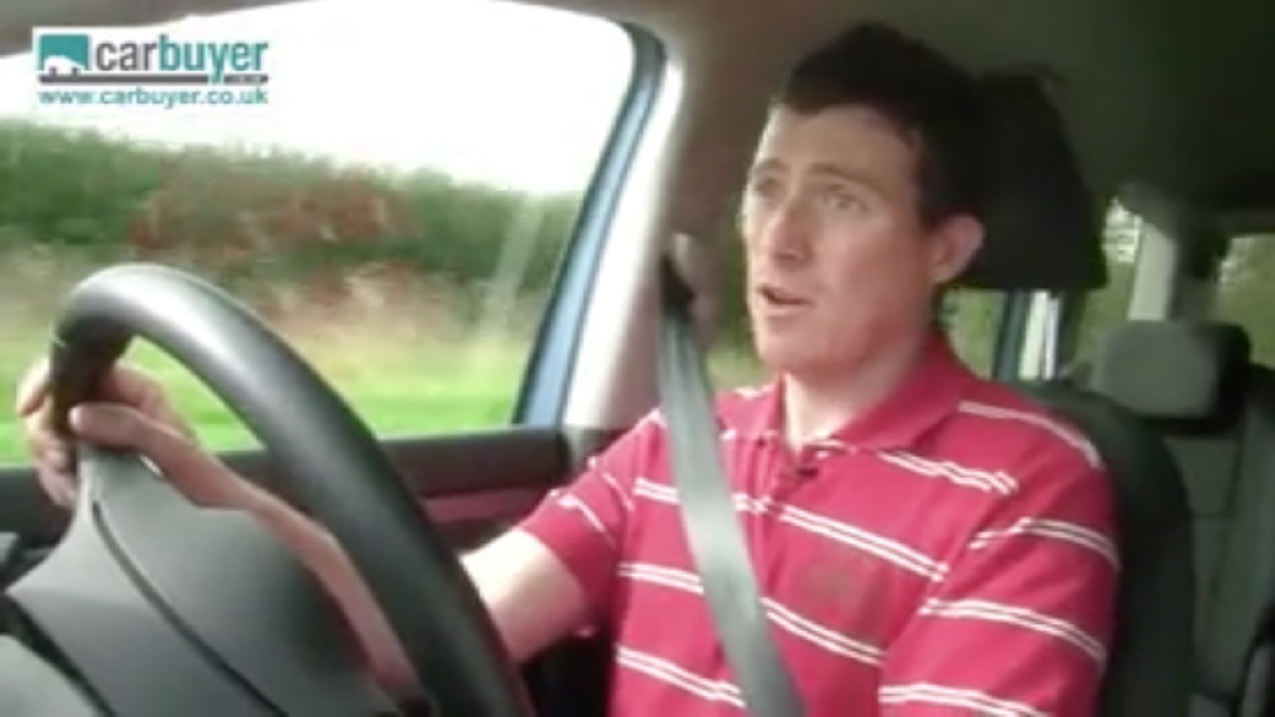}
  \end{subfigure}\hfill
  \begin{subfigure}{0.157\textwidth}\hspace{0.01\textwidth}
    \includegraphics[width=\linewidth]{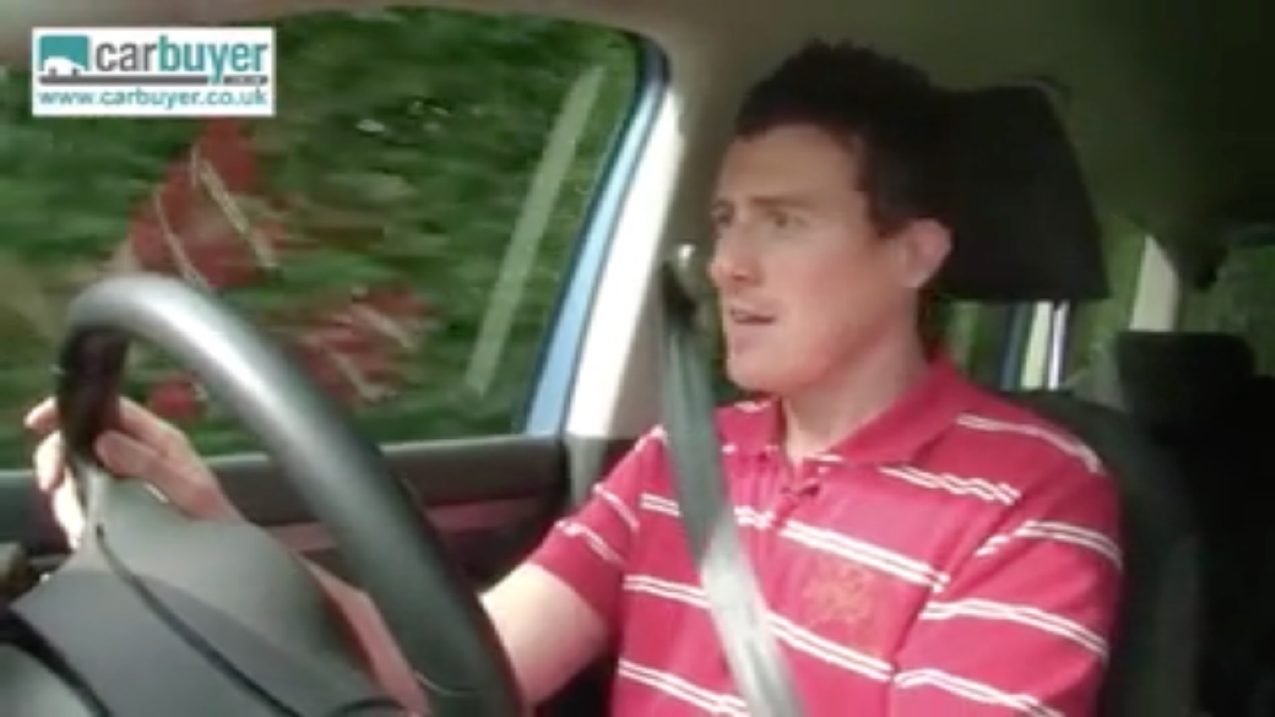}
  \end{subfigure}

  \vspace{0.3ex} 
  \begin{flushleft}
  \small
  \textbf{Ground truth:} a man wearing a red shirt that is driving a car.\\
  \textbf{+Adapter:} a man in a red shirt is driving a car.\\
  \textbf{+Q-Adapter:} a man in a \underline{red and white striped} shirt is driving a car.
  \end{flushleft}
  \vspace{1.0ex} 

  \begin{subfigure}{0.157\textwidth}\hspace{0.01\textwidth}
    \includegraphics[width=\linewidth]{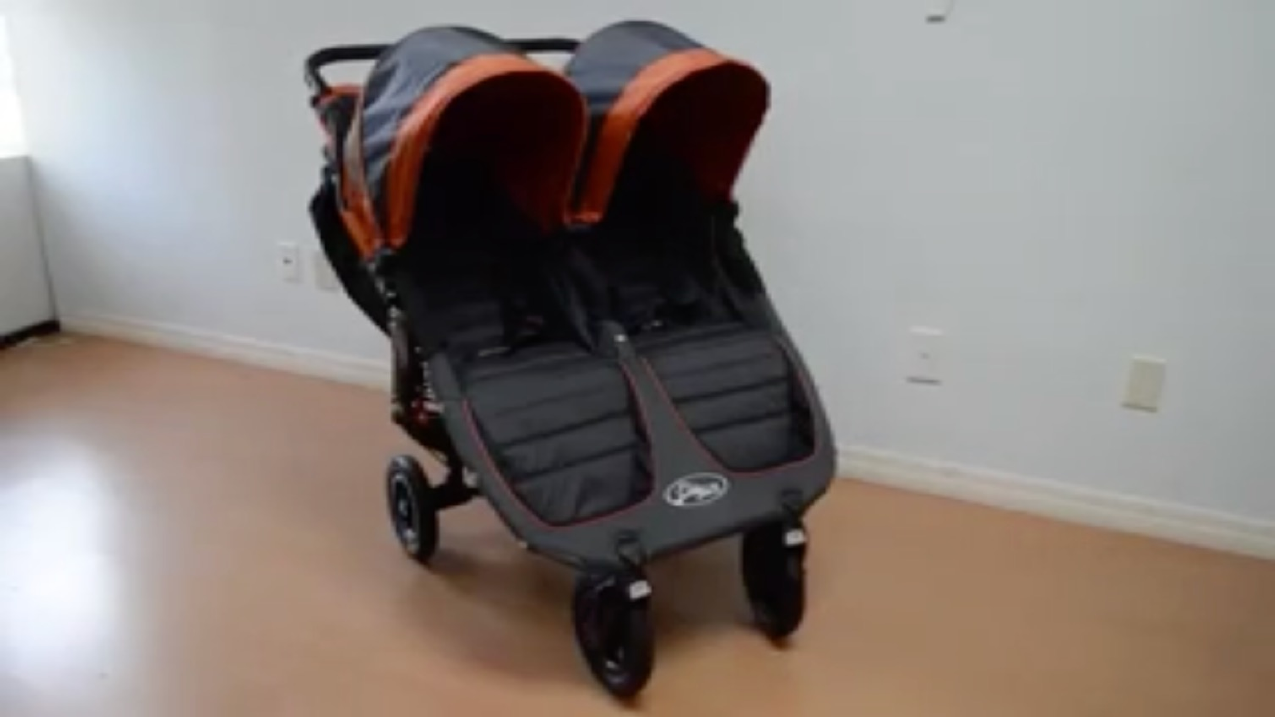}
  \end{subfigure}\hfill
  \begin{subfigure}{0.157\textwidth}\hspace{0.01\textwidth}
    \includegraphics[width=\linewidth]{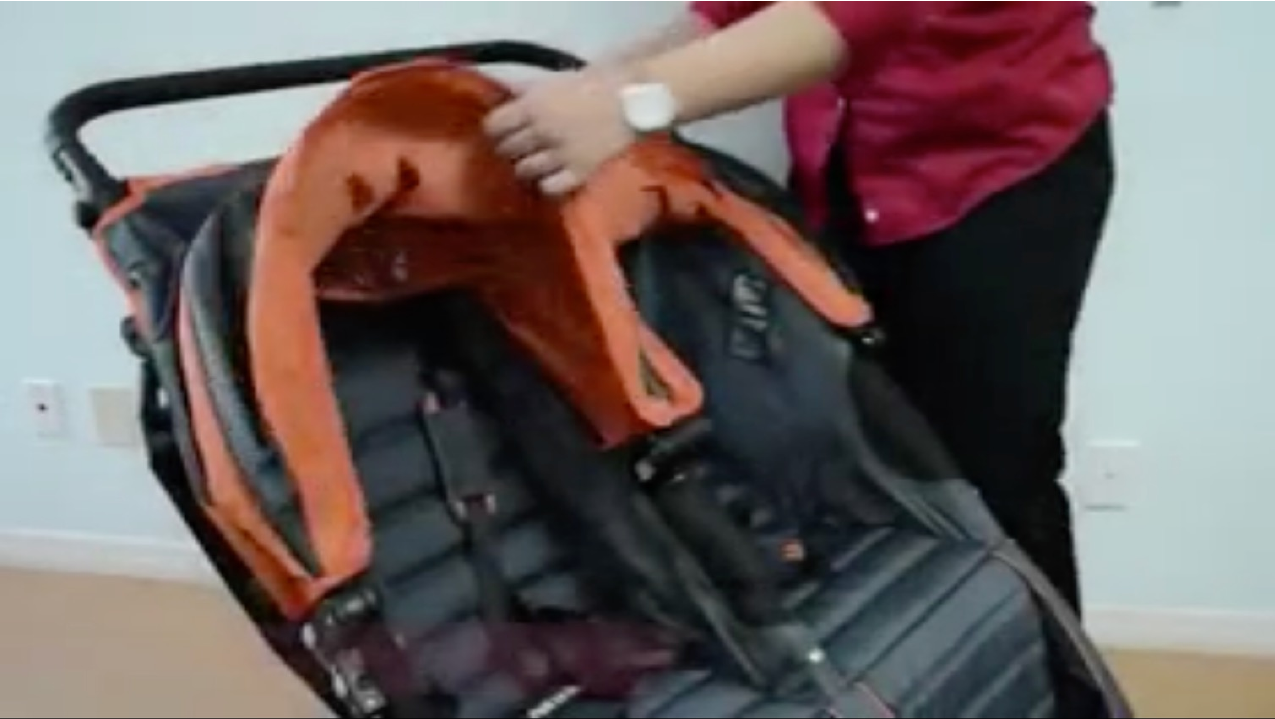}
  \end{subfigure}\hfill
  \begin{subfigure}{0.157\textwidth}\hspace{0.01\textwidth}
    \includegraphics[width=\linewidth]{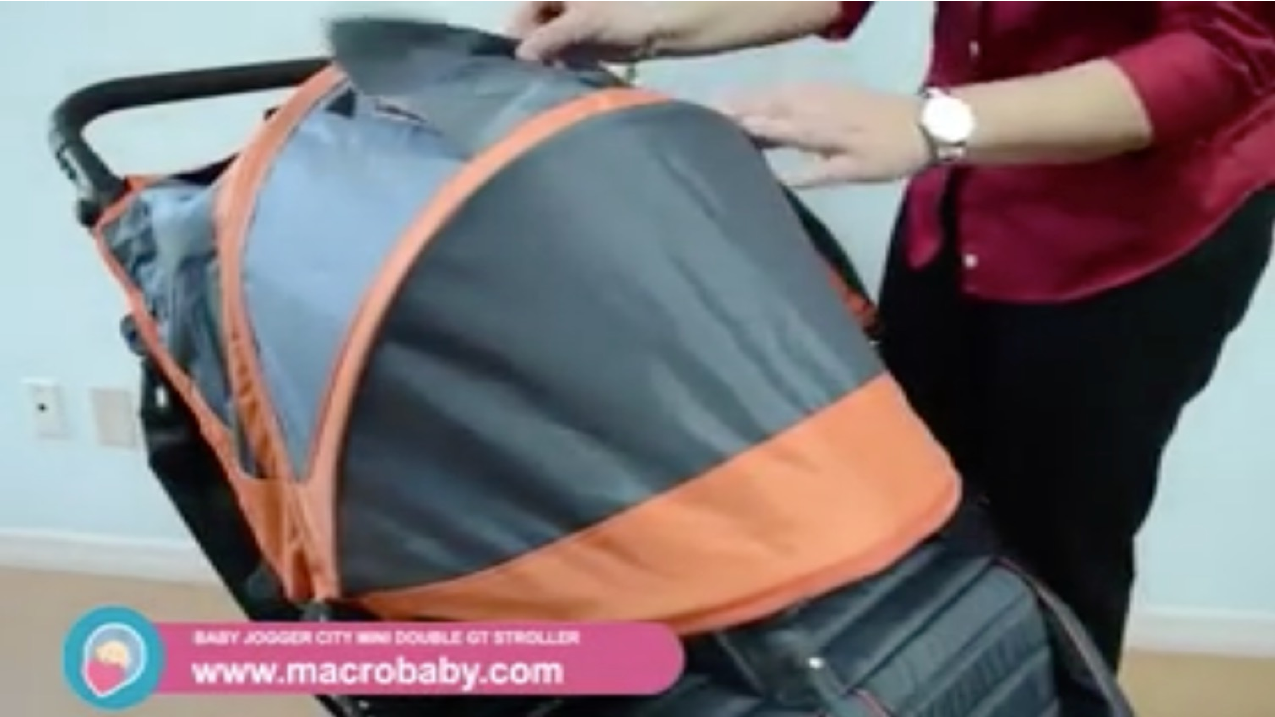}
  \end{subfigure}
  
  \vspace{0.3ex} 
  \begin{flushleft}
  \small
  \textbf{Ground truth:} a woman is adjusting the stroller.\\
  \textbf{+Adapter:} a \textcolor{red}{man} is demonstrating how to use a stroller.\\
  \textbf{+Q-Adapter:} a woman is demonstrating how to \underline{fold} a stroller.
  \end{flushleft}
  \vspace{1.0ex} 

  \begin{subfigure}{0.157\textwidth}\hspace{0.01\textwidth}
    \includegraphics[width=\linewidth]{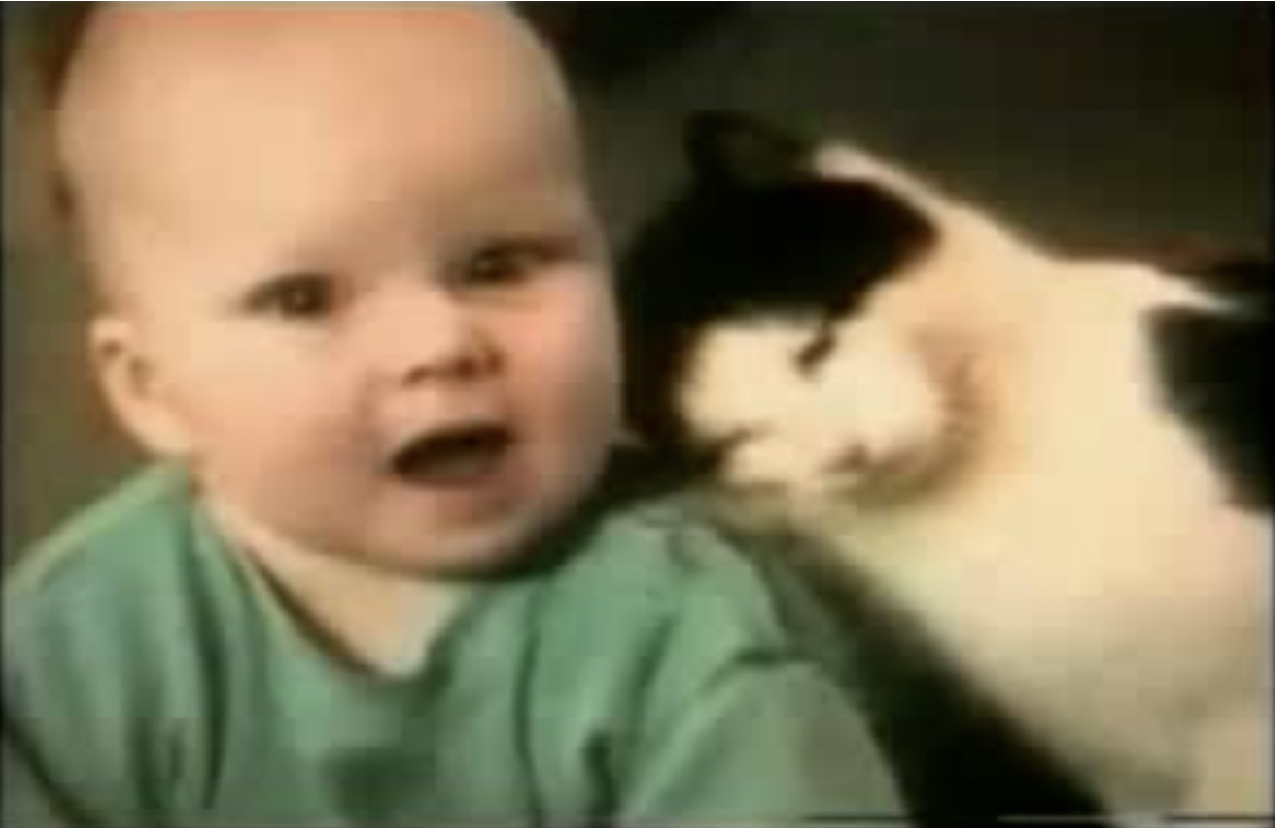}
  \end{subfigure}\hfill
  \begin{subfigure}{0.157\textwidth}\hspace{0.01\textwidth}
    \includegraphics[width=\linewidth]{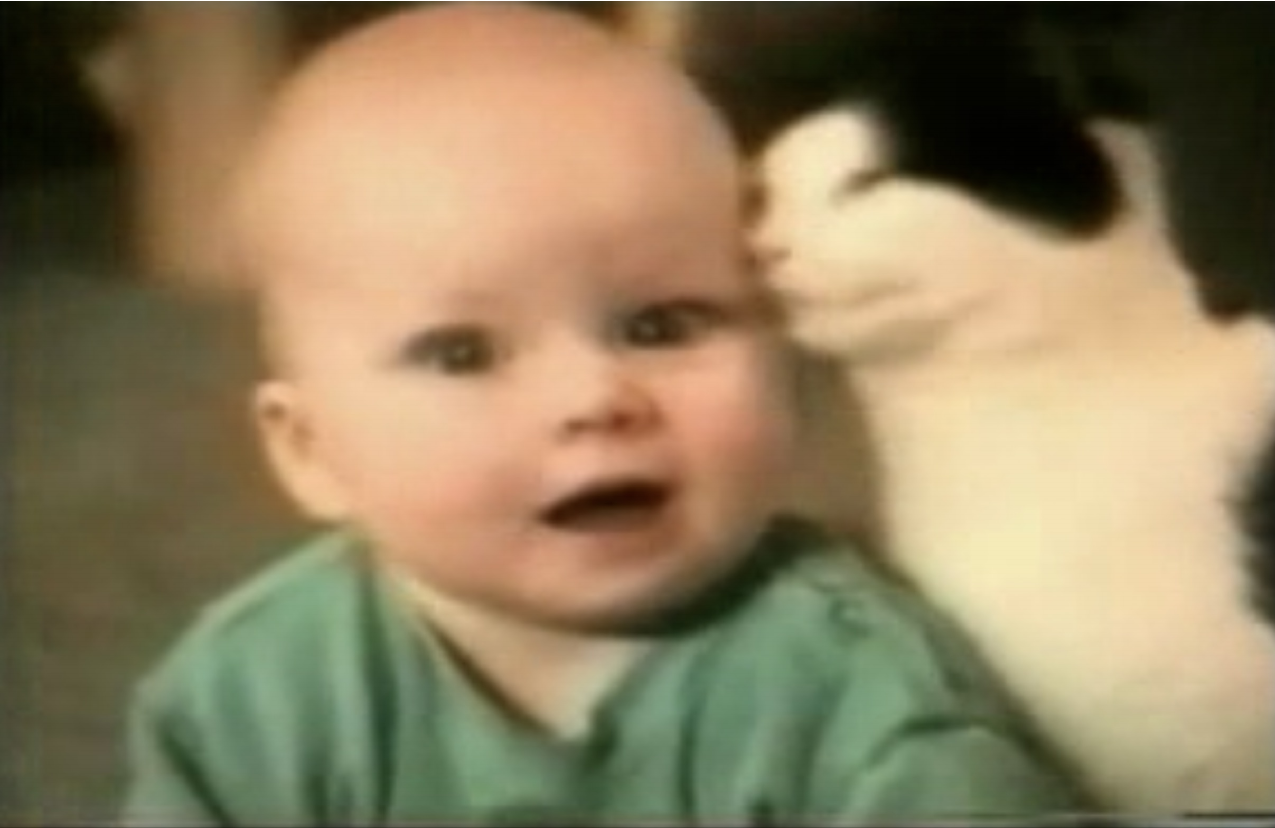}
  \end{subfigure}\hfill
  \begin{subfigure}{0.157\textwidth}\hspace{0.01\textwidth}
    \includegraphics[width=\linewidth]{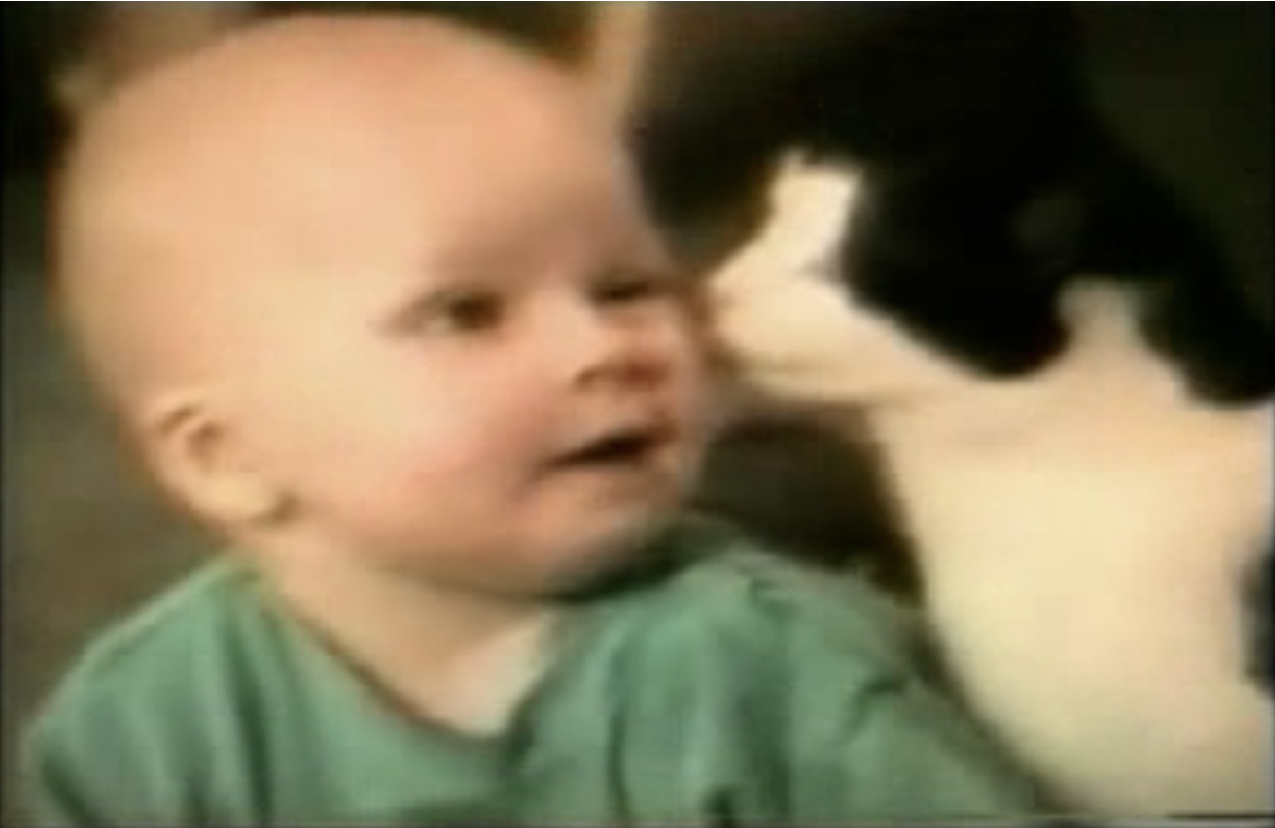}
  \end{subfigure}

  \vspace{0.3ex} 
  \begin{flushleft}
  \small
  \textbf{Ground truth:} a cat is rubbing against baby's face.\\
  \textbf{+Adapter:} a cat is licking a baby.\\
  \textbf{+Q-Adapter:} a cat is licking a baby's \underline{face}.
  \end{flushleft}
  \vspace{1.0ex} 
  
   \begin{subfigure}{0.157\textwidth}\hspace{0.01\textwidth}
    \includegraphics[width=\linewidth]{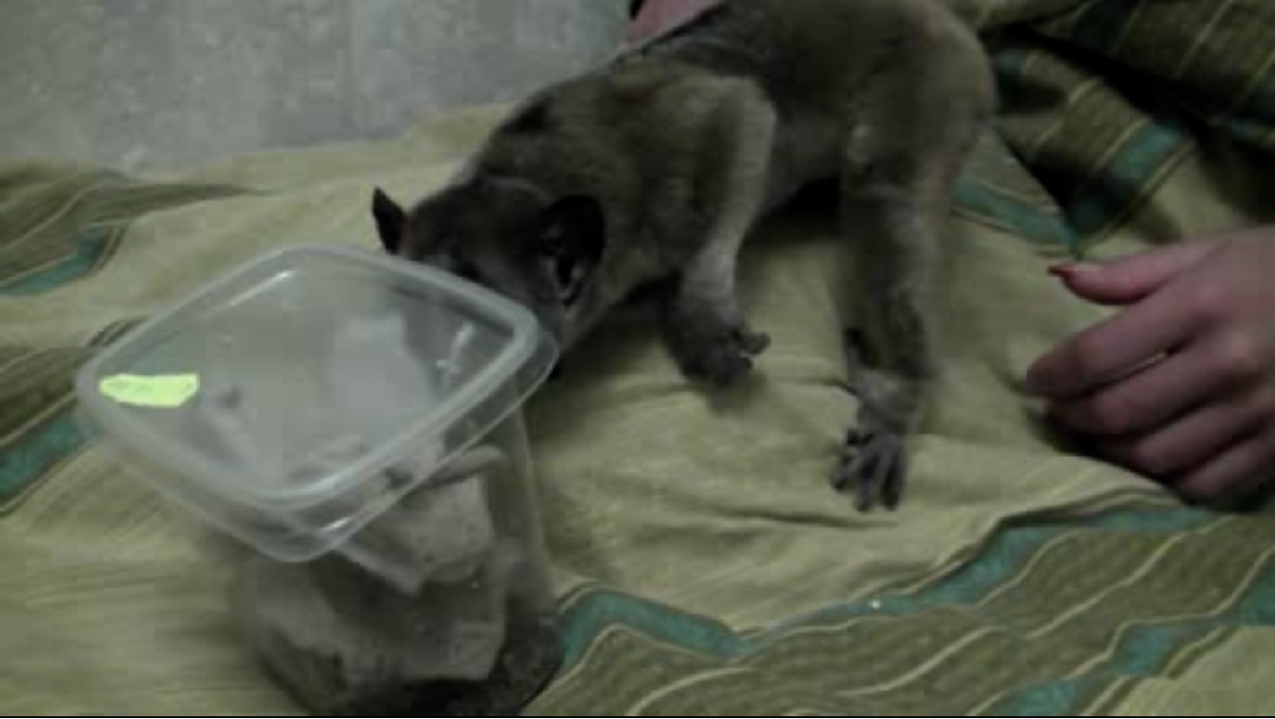}
  \end{subfigure}\hfill
  \begin{subfigure}{0.157\textwidth}\hspace{0.01\textwidth}
    \includegraphics[width=\linewidth]{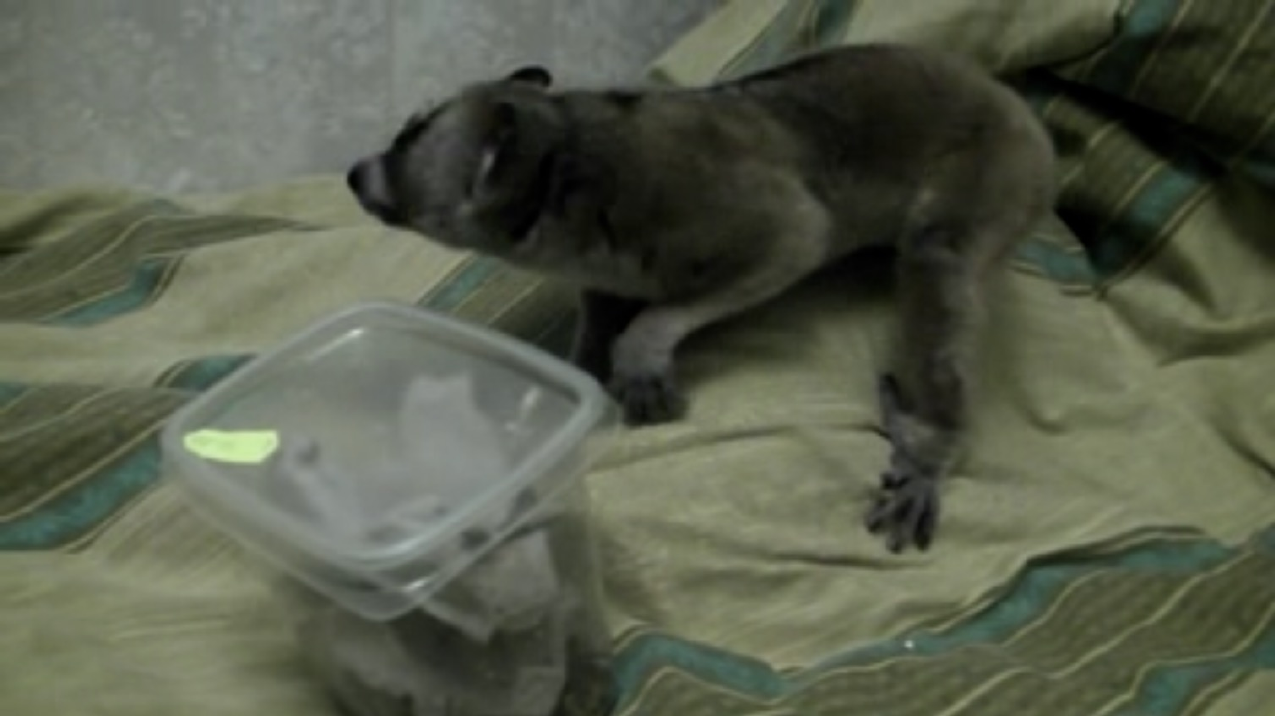}
  \end{subfigure}\hfill
  \begin{subfigure}{0.157\textwidth}\hspace{0.01\textwidth}
    \includegraphics[width=\linewidth]{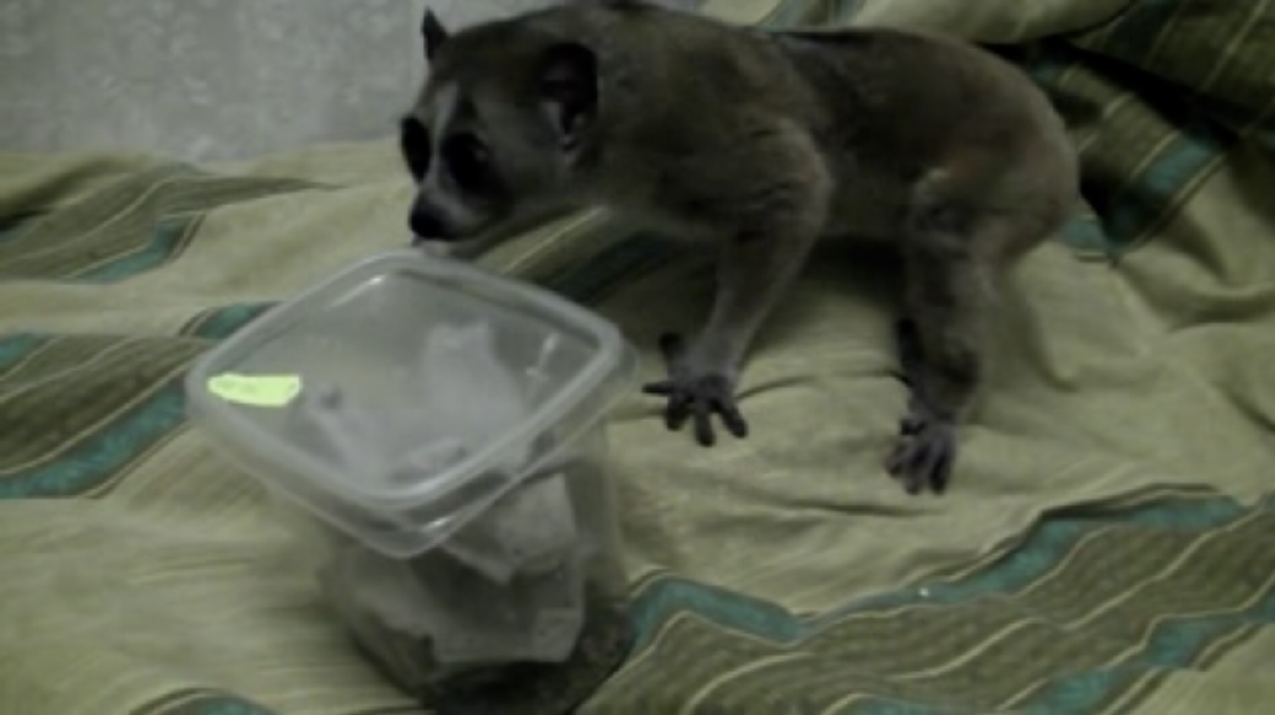}
  \end{subfigure}

  \vspace{0.3ex} 
  \begin{flushleft}
  \small
  \textbf{Ground truth:} an animal is inspecting a container.\\
  \textbf{+Adapter:} a cat is playing with a container.\\
  \textbf{+Q-Adapter:} a cat is playing with a \underline{plastic} container.
  \end{flushleft}
  \vspace{-10pt}
  \caption{Examples of generated captions from the MSR-VTT~\cite{xu2016msr} and MSVD~\cite{chen2011collecting} datasets.}
  \label{fig:msrvtt-example}
\end{figure}

Figure~\ref{fig:msrvtt-example} presents qualitative examples of captions generated from the MSR-VTT~\cite{xu2016msr} and MSVD~\cite{chen2011collecting} datasets. 
Compared to the baseline Adapter method~\cite{houlsby2019parameter}, the proposed Q-Adapter method generated captions that were more detailed, accurate, and context-aware.
In the first example, Q-Adapter correctly captured fine-grained visual details such as the ``red and white striped shirt'', indicating enhanced visual discrimination. 
In the second example, Q-Adapter correctly identified the subject as a ``woman'' and accurately described the action as ``folding'' the stroller, while Adapter misidentified the subject and provided a less specific description. 
Additional examples further highlight the ability of Q-Adapter to incorporate nuanced visual cues, such as distinguishing ``face'' and identifying object attributes like ``plastic container''.
These results qualitatively validated the effectiveness of Q-Adapter in producing rich and semantically precise video captions by attending to task-relevant visual features.

\begin{table}
\centering
\caption{Impact of the number of query tokens in the proposed Q-Adapter on the MSR-VTT~\cite{xu2016msr} dataset.
}
\label{tab:ablation_query_tokens}
\begin{tabular}{c|c|cccc}
\toprule
\#Queries & FT Ratio & B@4$\uparrow$ &  M$\uparrow$ &  R-L$\uparrow$ &  C$\uparrow$ \\
\midrule
  16  & 1.4\% &  49.84  &  \textbf{32.32}  &  \textbf{62.69}   & \textbf{67.38}   \\
  64  & 1.4\% &  \textbf{50.03}   & \underline{32.28}  & \underline{62.41}    & 66.46 \\
  112 & 1.5\% &  \underline{49.86}  & 31.95   &  62.17   & 66.88 \\
  224  & 1.6\% &  49.29  & \underline{32.28}   &  62.32   & \underline{67.14} \\
\bottomrule
\end{tabular}
\end{table}

\begin{table}
\centering
\caption{Effect of the proposed Q-Adapter placement in the Vision Encoder on the MSR-VTT~\cite{xu2016msr} dataset.
}
\label{tab:ablation_placement}
{
\begin{tabular}{l|c|cccc}
\toprule
Placement & FT Ratio &  B@4$\uparrow$ &  M$\uparrow$ &  R-L$\uparrow$ &  C$\uparrow$ \\
\midrule
Sequential & 1.4\% & 47.48  & 31.57   & 61.44   &  64.68 \\
Parallel-MLP & 0.7\% & \textbf{49.96}  & \underline{31.87}  &  \underline{62.37} & \underline{66.71} \\
Proposed &  1.4\% & \underline{49.84}  &  \textbf{32.32}  & \textbf{62.69}     & \textbf{67.38} \\
\bottomrule
\end{tabular}
}
\end{table}

\section{Conclusion}
\label{sec:conclusion}
We proposed Q-Adapter, a PEFT method designed to adapt Vision Encoder of an MLLM for the video captioning task. 
By inserting lightweight query-attention modules into Vision Encoder, it achieved effective adaptation with only 1.4\% of trainable parameters compared to the full fine-tuning approach. Extensive experiments on standard video captioning datasets, MSR-VTT~\cite{xu2016msr} and MSVD~\cite{chen2011collecting}, demonstrated that Q-Adapter achieved the state-of-the-art performance among methods that took the PEFT approach and remained competitive with the full fine-tuning approach. 
Through detailed ablation studies, we further analyzed the effects of query token count and adapter placement within Vision Encoder. 
The results showed that Q-Adapter struck an effective balance between caption quality and parameter efficiency, while maintaining flexibility for scalable integration into an existing MLLM.

In future work, we plan to explore the use of alternative pretrained backbones to assess the generalizability of Q-Adapter across diverse model architectures. 
We also aim to enhance query supervision and investigate strategies such as incorporating auxiliary knowledge via retrieval-augmented generation, as well as leveraging weakly supervised and self-supervised learning to better exploit the potential of pretrained MLLMs. 

\begin{acks}
This work was partly supported by Japan Society for the Promotion of Science (JSPS) KAKENHI 23K24868 and 25K00161.
\end{acks}

\bibliographystyle{ACM-Reference-Format}
\bibliography{references}


\begin{thebibliography}{45}


\ifx \showCODEN    \undefined \def \showCODEN     #1{\unskip}     \fi
\ifx \showDOI      \undefined \def \showDOI       #1{#1}\fi
\ifx \showISBNx    \undefined \def \showISBNx     #1{\unskip}     \fi
\ifx \showISBNxiii \undefined \def \showISBNxiii  #1{\unskip}     \fi
\ifx \showISSN     \undefined \def \showISSN      #1{\unskip}     \fi
\ifx \showLCCN     \undefined \def \showLCCN      #1{\unskip}     \fi
\ifx \shownote     \undefined \def \shownote      #1{#1}          \fi
\ifx \showarticletitle \undefined \def \showarticletitle #1{#1}   \fi
\ifx \showURL      \undefined \def \showURL       {\relax}        \fi
\providecommand\bibfield[2]{#2}
\providecommand\bibinfo[2]{#2}
\providecommand\natexlab[1]{#1}
\providecommand\showeprint[2][]{arXiv:#2}

\bibitem[Bai et~al\mbox{.}(2025)]%
        {qwen2.5-VL}
\bibfield{author}{\bibinfo{person}{Shuai Bai}, \bibinfo{person}{Keqin Chen}, \bibinfo{person}{Xuejing Liu}, \bibinfo{person}{Jialin Wang}, \bibinfo{person}{Wenbin Ge}, \bibinfo{person}{Sibo Song}, \bibinfo{person}{Kai Dang}, \bibinfo{person}{Peng Wang}, \bibinfo{person}{Shijie Wang}, \bibinfo{person}{Jun Tang}, \bibinfo{person}{Humen Zhong}, \bibinfo{person}{Yuanzhi Zhu}, \bibinfo{person}{Mingkun Yang}, \bibinfo{person}{Zhaohai Li}, \bibinfo{person}{Jianqiang Wan}, \bibinfo{person}{Pengfei Wang}, \bibinfo{person}{Wei Ding}, \bibinfo{person}{Zheren Fu}, \bibinfo{person}{Yiheng Xu}, \bibinfo{person}{Jiabo Ye}, \bibinfo{person}{Xi Zhang}, \bibinfo{person}{Tianbao Xie}, \bibinfo{person}{Zesen Cheng}, \bibinfo{person}{Hang Zhang}, \bibinfo{person}{Zhibo Yang}, \bibinfo{person}{Haiyang Xu}, {and} \bibinfo{person}{Junyang Lin}.} \bibinfo{year}{2025}\natexlab{}.
\newblock \showarticletitle{Qwen2.5-VL Technical Report}.
\newblock \bibinfo{journal}{\emph{Computing Research Repository arXiv Preprint, \textnormal{arXiv:2502.13923}}} (\bibinfo{year}{2025}).
\newblock


\bibitem[Banerjee and Lavie(2005)]%
        {banerjee2005meteor}
\bibfield{author}{\bibinfo{person}{Satanjeev Banerjee} {and} \bibinfo{person}{Alon Lavie}.} \bibinfo{year}{2005}\natexlab{}.
\newblock \showarticletitle{METEOR: An automatic metric for MT evaluation with improved correlation with human judgments}. In \bibinfo{booktitle}{\emph{Proc. ACL2005 Workshop on Intrinsic and Extrinsic Evaluation Measures for Machine Translation and/or Summarization}}. \bibinfo{pages}{65--72}.
\newblock


\bibitem[Caba~Heilbron et~al\mbox{.}(2015)]%
        {caba2015activitynet}
\bibfield{author}{\bibinfo{person}{Fabian Caba~Heilbron}, \bibinfo{person}{Victor Escorcia}, \bibinfo{person}{Bernard Ghanem}, {and} \bibinfo{person}{Juan Carlos~Niebles}.} \bibinfo{year}{2015}\natexlab{}.
\newblock \showarticletitle{ActivityNet: A large-scale video benchmark for human activity understanding}. In \bibinfo{booktitle}{\emph{Proc. 2015 {IEEE} Conference on Computer Vision and Pattern Recognition}}. \bibinfo{pages}{961--970}.
\newblock


\bibitem[Chen and Dolan(2011)]%
        {chen2011collecting}
\bibfield{author}{\bibinfo{person}{David~L Chen} {and} \bibinfo{person}{William~B Dolan}.} \bibinfo{year}{2011}\natexlab{}.
\newblock \showarticletitle{Collecting highly parallel data for paraphrase evaluation}. In \bibinfo{booktitle}{\emph{Proc. 49th Annual Meeting of the Association for Computational Linguistics: Human Language Technologies}}. \bibinfo{pages}{190--200}.
\newblock


\bibitem[Chen et~al\mbox{.}(2022)]%
        {chen2022adaptformer}
\bibfield{author}{\bibinfo{person}{Shoufa Chen}, \bibinfo{person}{Chongjian Ge}, \bibinfo{person}{Zhan Tong}, \bibinfo{person}{Jiangliu Wang}, \bibinfo{person}{Yibing Song}, \bibinfo{person}{Jue Wang}, {and} \bibinfo{person}{Ping Luo}.} \bibinfo{year}{2022}\natexlab{}.
\newblock \showarticletitle{AdaptFormer: Adapting vision Transformers for scalable visual recognition}.
\newblock \bibinfo{journal}{\emph{Advances in Neural Information Processing Systems}}  \bibinfo{volume}{35} (\bibinfo{year}{2022}), \bibinfo{pages}{16664--16678}.
\newblock


\bibitem[Chen et~al\mbox{.}(2023)]%
        {chen2023vast}
\bibfield{author}{\bibinfo{person}{Sihan Chen}, \bibinfo{person}{Handong Li}, \bibinfo{person}{Qunbo Wang}, \bibinfo{person}{Zijia Zhao}, \bibinfo{person}{Mingzhen Sun}, \bibinfo{person}{Xinxin Zhu}, {and} \bibinfo{person}{Jing Liu}.} \bibinfo{year}{2023}\natexlab{}.
\newblock \showarticletitle{VAST: A Vision-Audio-Subtitle-Text omni-modality foundation model and dataset}.
\newblock \bibinfo{journal}{\emph{Advances in Neural Information Processing Systems}}  \bibinfo{volume}{36} (\bibinfo{year}{2023}), \bibinfo{pages}{72842--72866}.
\newblock


\bibitem[Dosovitskiy et~al\mbox{.}(2020)]%
        {vit2020}
\bibfield{author}{\bibinfo{person}{Alexey Dosovitskiy}, \bibinfo{person}{Lucas Beyer}, \bibinfo{person}{Alexander Kolesnikov}, \bibinfo{person}{Dirk Weissenborn}, \bibinfo{person}{Xiaohua Zhai}, \bibinfo{person}{Thomas Unterthiner}, \bibinfo{person}{Mostafa Dehghani}, \bibinfo{person}{Matthias Minderer}, \bibinfo{person}{Georg Heigold}, \bibinfo{person}{Sylvain Gelly}, {et~al\mbox{.}}} \bibinfo{year}{2020}\natexlab{}.
\newblock \showarticletitle{An image is worth 16x16 words: Transformers for image recognition at scale}.
\newblock \bibinfo{journal}{\emph{Computing Research Repository arXiv Preprint, \textnormal{arXiv:2010.11929}}} (\bibinfo{year}{2020}).
\newblock


\bibitem[Floridi and Chiriatti(2020)]%
        {floridi2020gpt}
\bibfield{author}{\bibinfo{person}{Luciano Floridi} {and} \bibinfo{person}{Massimo Chiriatti}.} \bibinfo{year}{2020}\natexlab{}.
\newblock \showarticletitle{GPT-3: Its nature, scope, limits, and consequences}.
\newblock \bibinfo{journal}{\emph{Minds and Machines}}  \bibinfo{volume}{30} (\bibinfo{year}{2020}), \bibinfo{pages}{681--694}.
\newblock


\bibitem[Houlsby et~al\mbox{.}(2019)]%
        {houlsby2019parameter}
\bibfield{author}{\bibinfo{person}{Neil Houlsby}, \bibinfo{person}{Andrei Giurgiu}, \bibinfo{person}{Stanislaw Jastrzebski}, \bibinfo{person}{Bruna Morrone}, \bibinfo{person}{Quentin De~Laroussilhe}, \bibinfo{person}{Andrea Gesmundo}, \bibinfo{person}{Mona Attariyan}, {and} \bibinfo{person}{Sylvain Gelly}.} \bibinfo{year}{2019}\natexlab{}.
\newblock \showarticletitle{Parameter-efficient transfer learning for NLP}. In \bibinfo{booktitle}{\emph{Proc. 36th International Conference on Machine Learning}}. \bibinfo{pages}{2790--2799}.
\newblock


\bibitem[Hu et~al\mbox{.}(2022)]%
        {hu2022lora}
\bibfield{author}{\bibinfo{person}{Edward~J Hu}, \bibinfo{person}{Yelong Shen}, \bibinfo{person}{Phillip Wallis}, \bibinfo{person}{Zeyuan Allen-Zhu}, \bibinfo{person}{Yuanzhi Li}, \bibinfo{person}{Shean Wang}, \bibinfo{person}{Lu Wang}, {and} \bibinfo{person}{Weizhu Chen}.} \bibinfo{year}{2022}\natexlab{}.
\newblock \showarticletitle{LoRA: Low-Rank Adaptation of large language models}. In \bibinfo{booktitle}{\emph{Proc. 10th International Conference on Learning Representations}}. \bibinfo{pages}{3--16}.
\newblock


\bibitem[Huang et~al\mbox{.}(2024)]%
        {huang2024vtimellm}
\bibfield{author}{\bibinfo{person}{Bin Huang}, \bibinfo{person}{Xin Wang}, \bibinfo{person}{Hong Chen}, \bibinfo{person}{Zihan Song}, {and} \bibinfo{person}{Wenwu Zhu}.} \bibinfo{year}{2024}\natexlab{}.
\newblock \showarticletitle{VTimeLLM: Empower LLM to grasp video moments}. In \bibinfo{booktitle}{\emph{Proc. 2024 IEEE/CVF Conference on Computer Vision and Pattern Recognition}}. \bibinfo{pages}{14271--14280}.
\newblock


\bibitem[Kim et~al\mbox{.}(2024)]%
        {0Do}
\bibfield{author}{\bibinfo{person}{Minkuk Kim}, \bibinfo{person}{Hyeon~Bae Kim}, \bibinfo{person}{Jinyoung Moon}, \bibinfo{person}{Jinwoo Choi}, {and} \bibinfo{person}{Seong~Tae Kim}.} \bibinfo{year}{2024}\natexlab{}.
\newblock \showarticletitle{Do you remember? Dense video captioning with cross-modal memory retrieval}. In \bibinfo{booktitle}{\emph{Proc. 2024 IEEE/CVF Conference on Computer Vision and Pattern Recognition}}. \bibinfo{pages}{13894--13904}.
\newblock


\bibitem[Li et~al\mbox{.}(2021b)]%
        {2021Open}
\bibfield{author}{\bibinfo{person}{Bing Li}, \bibinfo{person}{Chunfeng Yuan}, \bibinfo{person}{Weiming Hu}, \bibinfo{person}{Ying Deng}, \bibinfo{person}{Ying Shan}, \bibinfo{person}{Zhongang Qi}, {and} \bibinfo{person}{Ziqi Zhang}.} \bibinfo{year}{2021}\natexlab{b}.
\newblock \showarticletitle{Open-book video captioning with retrieve-copy-generate network}. In \bibinfo{booktitle}{\emph{Proc. 2021 IEEE/CVF Conference on Computer Vision and Pattern Recognition}}. \bibinfo{pages}{9837--9846}.
\newblock


\bibitem[Li et~al\mbox{.}(2023b)]%
        {li2023blip}
\bibfield{author}{\bibinfo{person}{Junnan Li}, \bibinfo{person}{Dongxu Li}, \bibinfo{person}{Silvio Savarese}, {and} \bibinfo{person}{Steven Hoi}.} \bibinfo{year}{2023}\natexlab{b}.
\newblock \showarticletitle{{BLIP-2:} Bootstrapping Language-Image Pre-training with frozen image encoders and large language models}. In \bibinfo{booktitle}{\emph{Proc. 40th International Conference on Machine Learning}}. \bibinfo{pages}{19730--19742}.
\newblock


\bibitem[Li et~al\mbox{.}(2021a)]%
        {li2021align}
\bibfield{author}{\bibinfo{person}{Junnan Li}, \bibinfo{person}{Ramprasaath Selvaraju}, \bibinfo{person}{Akhilesh Gotmare}, \bibinfo{person}{Shafiq Joty}, \bibinfo{person}{Caiming Xiong}, {and} \bibinfo{person}{Steven Chu~Hong Hoi}.} \bibinfo{year}{2021}\natexlab{a}.
\newblock \showarticletitle{Align before fuse: Vision and language representation learning with momentum distillation}.
\newblock \bibinfo{journal}{\emph{Advances in Neural Information Processing Systems}}  \bibinfo{volume}{34} (\bibinfo{year}{2021}), \bibinfo{pages}{9694--9705}.
\newblock


\bibitem[Li et~al\mbox{.}(2023a)]%
        {li2024videochatchatcentricvideounderstanding}
\bibfield{author}{\bibinfo{person}{Kunchang Li}, \bibinfo{person}{Yinan He}, \bibinfo{person}{Yi Wang}, \bibinfo{person}{Yizhuo Li}, \bibinfo{person}{Wenhai Wang}, \bibinfo{person}{Ping Luo}, \bibinfo{person}{Yali Wang}, \bibinfo{person}{Limin Wang}, {and} \bibinfo{person}{Yu Qiao}.} \bibinfo{year}{2023}\natexlab{a}.
\newblock \showarticletitle{VideoChat: Chat-centric video understanding}.
\newblock \bibinfo{journal}{\emph{Computing Research Repository arXiv Preprint, \textnormal{arXiv:2305.06355}}} (\bibinfo{year}{2023}).
\newblock


\bibitem[Lin(2004)]%
        {lin2004rouge}
\bibfield{author}{\bibinfo{person}{Chin-Yew Lin}.} \bibinfo{year}{2004}\natexlab{}.
\newblock \showarticletitle{ROUGE: A package for automatic evaluation of summaries}. In \bibinfo{booktitle}{\emph{Proc. ACL2004 Workshop on Text Summarization Branches Out}}. \bibinfo{pages}{74--81}.
\newblock


\bibitem[Lin et~al\mbox{.}(2022)]%
        {lin2022swinbert}
\bibfield{author}{\bibinfo{person}{Jing Lin}, \bibinfo{person}{Linjie Cheng}, \bibinfo{person}{Zhe Gan}, \bibinfo{person}{Liwei Wang}, {and} \bibinfo{person}{Zicheng Liu}.} \bibinfo{year}{2022}\natexlab{}.
\newblock \showarticletitle{SwinBERT: End-to-end Transformers with sparse attention for video captioning}. In \bibinfo{booktitle}{\emph{Proc. 2022 IEEE/CVF Conference on Computer Vision and Pattern Recognition}}. \bibinfo{pages}{19373--19383}.
\newblock


\bibitem[Liu et~al\mbox{.}(2021)]%
        {liu2021swinT}
\bibfield{author}{\bibinfo{person}{Ze Liu}, \bibinfo{person}{Yutong Lin}, \bibinfo{person}{Yue Cao}, \bibinfo{person}{Han Hu}, \bibinfo{person}{Yixuan Wei}, \bibinfo{person}{Zheng Zhang}, \bibinfo{person}{Stephen Lin}, {and} \bibinfo{person}{Baining Guo}.} \bibinfo{year}{2021}\natexlab{}.
\newblock \showarticletitle{Swin Transformer: Hierarchical vision Transformer using shifted windows}. In \bibinfo{booktitle}{\emph{Proc. 18th IEEE/CVF International Conference on Computer Vision}}. \bibinfo{pages}{10012--10022}.
\newblock


\bibitem[Loshchilov and Hutter(2017)]%
        {loshchilov2017decoupled}
\bibfield{author}{\bibinfo{person}{Ilya Loshchilov} {and} \bibinfo{person}{Frank Hutter}.} \bibinfo{year}{2017}\natexlab{}.
\newblock \showarticletitle{Decoupled weight decay regularization}.
\newblock \bibinfo{journal}{\emph{Computing Research Repository arXiv Preprint, \textnormal{arXiv:1711.05101}}} (\bibinfo{year}{2017}).
\newblock


\bibitem[Luo et~al\mbox{.}(2022)]%
        {luo2022clip4clip}
\bibfield{author}{\bibinfo{person}{Huaishao Luo}, \bibinfo{person}{Lei Ji}, \bibinfo{person}{Ming Zhong}, \bibinfo{person}{Yang Chen}, \bibinfo{person}{Wen Lei}, \bibinfo{person}{Nan Duan}, {and} \bibinfo{person}{Tianrui Li}.} \bibinfo{year}{2022}\natexlab{}.
\newblock \showarticletitle{CLIP4Clip: An empirical study of CLIP for end to end video Clip retrieval and captioning}.
\newblock \bibinfo{journal}{\emph{Neurocomputing}}  \bibinfo{volume}{508} (\bibinfo{year}{2022}), \bibinfo{pages}{293--304}.
\newblock


\bibitem[Mercea et~al\mbox{.}(2024)]%
        {mercea2024time}
\bibfield{author}{\bibinfo{person}{Otniel-Bogdan Mercea}, \bibinfo{person}{Alexey Gritsenko}, \bibinfo{person}{Cordelia Schmid}, {and} \bibinfo{person}{Anurag Arnab}.} \bibinfo{year}{2024}\natexlab{}.
\newblock \showarticletitle{Time-memory-and parameter-efficient visual adaptation}. In \bibinfo{booktitle}{\emph{Proc. 2024 IEEE/CVF Conference on Computer Vision and Pattern Recognition}}. \bibinfo{pages}{5536--5545}.
\newblock


\bibitem[Pan et~al\mbox{.}(2020)]%
        {2020Spatio}
\bibfield{author}{\bibinfo{person}{Boxiao Pan}, \bibinfo{person}{Haoye Cai}, \bibinfo{person}{De~An Huang}, \bibinfo{person}{Kuan~Hui Lee}, \bibinfo{person}{Adrien Gaidon}, \bibinfo{person}{Ehsan Adeli}, {and} \bibinfo{person}{Juan~Carlos Niebles}.} \bibinfo{year}{2020}\natexlab{}.
\newblock \showarticletitle{Spatio-temporal graph for video captioning with knowledge distillation}. In \bibinfo{booktitle}{\emph{Proc. 2020 IEEE/CVF Conference on Computer Vision and Pattern Recognition}}. \bibinfo{pages}{10867--10876}.
\newblock


\bibitem[Pan et~al\mbox{.}(2016)]%
        {pan2016jointly}
\bibfield{author}{\bibinfo{person}{Yingwei Pan}, \bibinfo{person}{Tao Mei}, \bibinfo{person}{Ting Yao}, \bibinfo{person}{Houqiang Li}, {and} \bibinfo{person}{Yong Rui}.} \bibinfo{year}{2016}\natexlab{}.
\newblock \showarticletitle{Jointly modeling embedding and translation to bridge video and language}. In \bibinfo{booktitle}{\emph{Proc. 2016 {IEEE} Conference on Computer Vision and Pattern Recognition}}. \bibinfo{pages}{4594--4602}.
\newblock


\bibitem[Papineni et~al\mbox{.}(2002)]%
        {papineni2002bleu}
\bibfield{author}{\bibinfo{person}{Kishore Papineni}, \bibinfo{person}{Salim Roukos}, \bibinfo{person}{Todd Ward}, {and} \bibinfo{person}{Wei-Jing Zhu}.} \bibinfo{year}{2002}\natexlab{}.
\newblock \showarticletitle{BLEU: A method for automatic evaluation of machine translation}. In \bibinfo{booktitle}{\emph{Proc. 40th Annual Meeting of the Association for Computational Linguistics}}. \bibinfo{pages}{311--318}.
\newblock


\bibitem[Radford et~al\mbox{.}(2021)]%
        {radford2021learning}
\bibfield{author}{\bibinfo{person}{Alec Radford}, \bibinfo{person}{Jong~Wook Kim}, \bibinfo{person}{Chris Hallacy}, \bibinfo{person}{Aditya Ramesh}, \bibinfo{person}{Gabriel Goh}, \bibinfo{person}{Sandhini Agarwal}, \bibinfo{person}{Girish Sastry}, \bibinfo{person}{Amanda Askell}, \bibinfo{person}{Pamela Mishkin}, \bibinfo{person}{Jack Clark}, \bibinfo{person}{Gretchen Krueger}, {and} \bibinfo{person}{Ilya Sutskever}.} \bibinfo{year}{2021}\natexlab{}.
\newblock \showarticletitle{Learning transferable visual models from natural language supervision}. In \bibinfo{booktitle}{\emph{Proc. 38th International Conference on Machine Learning}}. \bibinfo{pages}{8748--8763}.
\newblock


\bibitem[Rohrbach et~al\mbox{.}(2015)]%
        {rohrbach2015dataset}
\bibfield{author}{\bibinfo{person}{Anna Rohrbach}, \bibinfo{person}{Marcus Rohrbach}, \bibinfo{person}{Niket Tandon}, {and} \bibinfo{person}{Bernt Schiele}.} \bibinfo{year}{2015}\natexlab{}.
\newblock \showarticletitle{A dataset for movie description}. In \bibinfo{booktitle}{\emph{Proc. 2015 {IEEE} Conference on Computer Vision and Pattern Recognition}}. \bibinfo{pages}{3202--3212}.
\newblock


\bibitem[Rohrbach et~al\mbox{.}(2013)]%
        {rohrbach2013translating}
\bibfield{author}{\bibinfo{person}{Marcus Rohrbach}, \bibinfo{person}{Wei Qiu}, \bibinfo{person}{Ivan Titov}, \bibinfo{person}{Stefan Thater}, \bibinfo{person}{Manfred Pinkal}, {and} \bibinfo{person}{Bernt Schiele}.} \bibinfo{year}{2013}\natexlab{}.
\newblock \showarticletitle{Translating video content to natural language descriptions}. In \bibinfo{booktitle}{\emph{Proc. 14th IEEE International Conference on Computer Vision}}. \bibinfo{pages}{433--440}.
\newblock


\bibitem[R{\"{u}}ckl{\'{e}} et~al\mbox{.}(2021)]%
        {ruckle2020adapterdrop}
\bibfield{author}{\bibinfo{person}{Andreas R{\"{u}}ckl{\'{e}}}, \bibinfo{person}{Gregor Geigle}, \bibinfo{person}{Max Glockner}, \bibinfo{person}{Tilman Beck}, \bibinfo{person}{Jonas Pfeiffer}, \bibinfo{person}{Nils Reimers}, {and} \bibinfo{person}{Iryna Gurevych}.} \bibinfo{year}{2021}\natexlab{}.
\newblock \showarticletitle{{AdapterDrop}: On the efficiency of adapters in Transformers}. In \bibinfo{booktitle}{\emph{Proc. 2021 Conference on Empirical Methods in Natural Language Processing}}. \bibinfo{pages}{7930--7946}.
\newblock


\bibitem[Seo et~al\mbox{.}(2022)]%
        {seo2022end}
\bibfield{author}{\bibinfo{person}{Paul~Hongsuck Seo}, \bibinfo{person}{Arsha Nagrani}, \bibinfo{person}{Anurag Arnab}, {and} \bibinfo{person}{Cordelia Schmid}.} \bibinfo{year}{2022}\natexlab{}.
\newblock \showarticletitle{End-to-end generative pretraining for multimodal video captioning}. In \bibinfo{booktitle}{\emph{Proc. 2022 IEEE/CVF Conference on Computer Vision and Pattern Recognition}}. \bibinfo{pages}{17959--17968}.
\newblock


\bibitem[Shi et~al\mbox{.}(2020)]%
        {2020Learning}
\bibfield{author}{\bibinfo{person}{Botian Shi}, \bibinfo{person}{Lei Ji}, \bibinfo{person}{Zhendong Niu}, \bibinfo{person}{Nan Duan}, \bibinfo{person}{Ming Zhou}, {and} \bibinfo{person}{Xilin Chen}.} \bibinfo{year}{2020}\natexlab{}.
\newblock \showarticletitle{Learning semantic concepts and temporal alignment for narrated video procedural captioning}. In \bibinfo{booktitle}{\emph{Proc. 28th {ACM} International Conference on Multimedia}}. \bibinfo{pages}{4355--4363}.
\newblock


\bibitem[Sung et~al\mbox{.}(2022)]%
        {sung2022vl}
\bibfield{author}{\bibinfo{person}{Yi-Lin Sung}, \bibinfo{person}{Jaemin Cho}, {and} \bibinfo{person}{Mohit Bansal}.} \bibinfo{year}{2022}\natexlab{}.
\newblock \showarticletitle{VL-Adapter: Parameter-efficient transfer learning for Vision-and-Language tasks}. In \bibinfo{booktitle}{\emph{Proc. 2022 IEEE/CVF Conference on Computer Vision and Pattern Recognition}}. \bibinfo{pages}{5227--5237}.
\newblock


\bibitem[Tang et~al\mbox{.}(2021)]%
        {tang2021clip4caption}
\bibfield{author}{\bibinfo{person}{Mingkang Tang}, \bibinfo{person}{Zhanyu Wang}, \bibinfo{person}{Zhenhua Liu}, \bibinfo{person}{Fengyun Rao}, \bibinfo{person}{Dian Li}, {and} \bibinfo{person}{Xiu Li}.} \bibinfo{year}{2021}\natexlab{}.
\newblock \showarticletitle{CLIP4Caption: CLIP for video Caption}. In \bibinfo{booktitle}{\emph{Proc. 29th ACM International Conference on Multimedia}}. \bibinfo{pages}{4858--4862}.
\newblock


\bibitem[Touvron et~al\mbox{.}(2023)]%
        {touvron2023llama}
\bibfield{author}{\bibinfo{person}{Hugo Touvron}, \bibinfo{person}{Louis Martin}, \bibinfo{person}{Kevin Stone}, \bibinfo{person}{Peter Albert}, \bibinfo{person}{Amjad Almahairi}, \bibinfo{person}{Yasmine Babaei}, \bibinfo{person}{Nikolay Bashlykov}, \bibinfo{person}{Soumya Batra}, \bibinfo{person}{Prajjwal Bhargava}, \bibinfo{person}{Shruti Bhosale}, {et~al\mbox{.}}} \bibinfo{year}{2023}\natexlab{}.
\newblock \showarticletitle{Llama 2: Open foundation and fine-tuned chat models}.
\newblock \bibinfo{journal}{\emph{Computing Research Repository arXiv Preprint, \textnormal{arXiv:2307.09288}}} (\bibinfo{year}{2023}).
\newblock


\bibitem[Vedantam et~al\mbox{.}(2015)]%
        {vedantam2015cider}
\bibfield{author}{\bibinfo{person}{Ramakrishna Vedantam}, \bibinfo{person}{C Lawrence~Zitnick}, {and} \bibinfo{person}{Devi Parikh}.} \bibinfo{year}{2015}\natexlab{}.
\newblock \showarticletitle{{CIDEr}: Consensus-based Image Description Evaluation}. In \bibinfo{booktitle}{\emph{Proc. 2015 {IEEE} Conference on Computer Vision and Pattern Recognition}}. \bibinfo{pages}{4566--4575}.
\newblock


\bibitem[Xu et~al\mbox{.}(2023)]%
        {xu2023mplug}
\bibfield{author}{\bibinfo{person}{Haiyang Xu}, \bibinfo{person}{Qinghao Ye}, \bibinfo{person}{Ming Yan}, \bibinfo{person}{Yaya Shi}, \bibinfo{person}{Jiabo Ye}, \bibinfo{person}{Yuanhong Xu}, \bibinfo{person}{Chenliang Li}, \bibinfo{person}{Bin Bi}, \bibinfo{person}{Qi Qian}, \bibinfo{person}{Wei Wang}, \bibinfo{person}{Guohai Xu}, \bibinfo{person}{Ji Zhang}, \bibinfo{person}{Songfang Huang}, \bibinfo{person}{Fei Huang}, {and} \bibinfo{person}{Jingren Zhou}.} \bibinfo{year}{2023}\natexlab{}.
\newblock \showarticletitle{{mPLUG-2}: {A} modularized multi-modal foundation model across text, image and video}. In \bibinfo{booktitle}{\emph{Proc. 40th International Conference on Machine Learning}}. \bibinfo{pages}{38728--38748}.
\newblock


\bibitem[Xu et~al\mbox{.}(2016)]%
        {xu2016msr}
\bibfield{author}{\bibinfo{person}{Jun Xu}, \bibinfo{person}{Tao Mei}, \bibinfo{person}{Ting Yao}, {and} \bibinfo{person}{Yong Rui}.} \bibinfo{year}{2016}\natexlab{}.
\newblock \showarticletitle{{MSR-VTT:} {A} large video description dataset for bridging video and language}. In \bibinfo{booktitle}{\emph{Proc. 2016 {IEEE} Conference on Computer Vision and Pattern Recognition}}. \bibinfo{pages}{5288--5296}.
\newblock


\bibitem[Yang et~al\mbox{.}(2023a)]%
        {yang2023vid2seq}
\bibfield{author}{\bibinfo{person}{Antoine Yang}, \bibinfo{person}{Arsha Nagrani}, \bibinfo{person}{Paul~Hongsuck Seo}, \bibinfo{person}{Antoine Miech}, \bibinfo{person}{Jordi Pont-Tuset}, \bibinfo{person}{Ivan Laptev}, \bibinfo{person}{Josef Sivic}, {and} \bibinfo{person}{Cordelia Schmid}.} \bibinfo{year}{2023}\natexlab{a}.
\newblock \showarticletitle{{Vid2Seq}: Large-scale pretraining of a visual language model for dense video captioning}. In \bibinfo{booktitle}{\emph{Proc. 2023 IEEE/CVF Conference on Computer Vision and Pattern Recognition}}. \bibinfo{pages}{10714--10726}.
\newblock


\bibitem[Yang et~al\mbox{.}(2023b)]%
        {yangaim}
\bibfield{author}{\bibinfo{person}{Taojiannan Yang}, \bibinfo{person}{Yi Zhu}, \bibinfo{person}{Yusheng Xie}, \bibinfo{person}{Aston Zhang}, \bibinfo{person}{Chen Chen}, {and} \bibinfo{person}{Mu Li}.} \bibinfo{year}{2023}\natexlab{b}.
\newblock \showarticletitle{AIM: Adapting Image Models for efficient video action recognition}. In \bibinfo{booktitle}{\emph{Proc. 11th International Conference on Learning Representations}}. \bibinfo{pages}{1--14}.
\newblock


\bibitem[Ye et~al\mbox{.}(2023)]%
        {ye2023hitea}
\bibfield{author}{\bibinfo{person}{Qinghao Ye}, \bibinfo{person}{Guohai Xu}, \bibinfo{person}{Ming Yan}, \bibinfo{person}{Haiyang Xu}, \bibinfo{person}{Qi Qian}, \bibinfo{person}{Ji Zhang}, {and} \bibinfo{person}{Fei Huang}.} \bibinfo{year}{2023}\natexlab{}.
\newblock \showarticletitle{{HiTeA}: Hierarchical Temporal-Aware video--language pre-training}. In \bibinfo{booktitle}{\emph{Proc. 19th IEEE/CVF International Conference on Computer Vision}}. \bibinfo{pages}{15405--15416}.
\newblock


\bibitem[Yin et~al\mbox{.}(2025)]%
        {yin20245}
\bibfield{author}{\bibinfo{person}{Dongshuo Yin}, \bibinfo{person}{Leiyi Hu}, \bibinfo{person}{Bin Li}, \bibinfo{person}{Youqun Zhang}, {and} \bibinfo{person}{Xue Yang}.} \bibinfo{year}{2025}\natexlab{}.
\newblock \showarticletitle{5\%>100\%: Breaking performance shackles of full fine-tuning on visual recognition tasks}. In \bibinfo{booktitle}{\emph{Proc. 2025 IEEE/CVF Conference on Computer Vision and Pattern Recognition}}. \bibinfo{pages}{20071--20081}.
\newblock


\bibitem[Yin et~al\mbox{.}(2023)]%
        {2024A}
\bibfield{author}{\bibinfo{person}{Shukang Yin}, \bibinfo{person}{Chaoyou Fu}, \bibinfo{person}{Sirui Zhao}, \bibinfo{person}{Ke Li}, \bibinfo{person}{Xing Sun}, \bibinfo{person}{Tong Xu}, {and} \bibinfo{person}{Enhong Chen}.} \bibinfo{year}{2023}\natexlab{}.
\newblock \showarticletitle{A survey on multimodal large language models}.
\newblock \bibinfo{journal}{\emph{Computing Research Repository arXiv Preprint, \textnormal{arXiv:2306.13549}}} (\bibinfo{year}{2023}).
\newblock


\bibitem[Yu et~al\mbox{.}(2024)]%
        {yu2024eliciting}
\bibfield{author}{\bibinfo{person}{Keunwoo Yu}, \bibinfo{person}{Zheyuan Zhang}, \bibinfo{person}{Fengyuan Hu}, \bibinfo{person}{Shane Storks}, {and} \bibinfo{person}{Joyce Chai}.} \bibinfo{year}{2024}\natexlab{}.
\newblock \showarticletitle{Eliciting in-context learning in vision-language models for videos through curated data distributional properties}. In \bibinfo{booktitle}{\emph{Proc. 2024 Conference on Empirical Methods in Natural Language Processing}}. \bibinfo{pages}{20416--20431}.
\newblock


\bibitem[Zhai et~al\mbox{.}(2023)]%
        {zhai2023investigating}
\bibfield{author}{\bibinfo{person}{Yuexiang Zhai}, \bibinfo{person}{Shengbang Tong}, \bibinfo{person}{Xiao Li}, \bibinfo{person}{Mu Cai}, \bibinfo{person}{Qing Qu}, \bibinfo{person}{Yong~Jae Lee}, {and} \bibinfo{person}{Yi Ma}.} \bibinfo{year}{2023}\natexlab{}.
\newblock \showarticletitle{Investigating the catastrophic forgetting in multimodal large language models}.
\newblock \bibinfo{journal}{\emph{Computing Research Repository arXiv Preprint, \textnormal{arXiv:2309.10313}}} (\bibinfo{year}{2023}).
\newblock


\bibitem[Zhang et~al\mbox{.}(2023)]%
        {zhang2023llama}
\bibfield{author}{\bibinfo{person}{Renrui Zhang}, \bibinfo{person}{Jiaming Han}, \bibinfo{person}{Chris Liu}, \bibinfo{person}{Peng Gao}, \bibinfo{person}{Aojun Zhou}, \bibinfo{person}{Xiangfei Hu}, \bibinfo{person}{Shilin Yan}, \bibinfo{person}{Pan Lu}, \bibinfo{person}{Hongsheng Li}, {and} \bibinfo{person}{Yu Qiao}.} \bibinfo{year}{2023}\natexlab{}.
\newblock \showarticletitle{LLaMA-Adapter: Efficient fine-tuning of language models with zero-init attention}.
\newblock \bibinfo{journal}{\emph{Computing Research Repository arXiv Preprint, \textnormal{arXiv preprint arXiv:2303.16199}}} (\bibinfo{year}{2023}).
\newblock


\end{thebibliography}

\end{document}